\documentclass{article}
\usepackage{iclr2027_conference,times}

\usepackage{graphicx}
\usepackage[table]{xcolor}
\usepackage{hyperref}
\usepackage{url}
\usepackage{caption}

\usepackage{amsmath}
\usepackage{amssymb}
\usepackage{booktabs}
\usepackage{float}
\usepackage{flafter}
\usepackage{multirow}
\usepackage{pifont}
\usepackage{siunitx}
\usepackage{tabularx}
\usepackage{wrapfig}

\definecolor{POSRbg}{HTML}{F3F7F9}
\definecolor{POSRaccent}{HTML}{2F6173}
\hypersetup{
    colorlinks=true,
    citecolor=POSRaccent,
    linkcolor=black,
    urlcolor=black,
    filecolor=black,
    pdfborder={0 0 0}
}
\newcolumntype{C}{>{\centering\arraybackslash}p{0.075\textwidth}}
\newcommand{\best}[1]{\multicolumn{1}{c}{\textbf{#1}}}
\newcommand{\second}[1]{\multicolumn{1}{c}{\underline{#1}}}
\newcommand{\gain}[1]{\textcolor{POSRaccent}{\textbf{#1}}}
\newcommand{\vismaintablestyle}{\footnotesize\renewcommand{\arraystretch}{1.10}}

\title{Learning to Reason with Persistent Object States for Video Instance Segmentation}
\author{\parbox[t]{\textwidth}{\raggedright\normalfont\normalsize
    \textbf{%
    Yongxue Xu$^{1,2,*}$, Boxue Yang$^{1,*}$, Ziqian Liu$^{2}$, Shaoqiu Zhang$^{1}$, Rui Qian$^{3}$, Haopeng Chen$^{1,\dagger}$}\\[5pt]
    $^1$Shanghai Jiao Tong University\quad
    $^2$Sun Yat-sen University\quad
    $^3$Fudan University\\[3pt]
    \small
    \href{mailto:xuyx85@mail2.sysu.edu.cn}{\texttt{xuyx85@mail2.sysu.edu.cn}}\quad
    \texttt{\{\href{mailto:yangboxue@sjtu.edu.cn}{yangboxue},
    \href{mailto:chen-hp@sjtu.edu.cn}{chen-hp}\}@sjtu.edu.cn}
}}

\iclrfinalcopy
\makeatletter
\renewcommand{\@maketitle}{\vbox{\hsize\textwidth
    {\LARGE\scshape\@title\par}
    \vskip8pt
    \@author\par
    \vskip2pt
}}
\makeatother
\renewenvironment{abstract}{\centerline{%
    \raisebox{-5pt}[\height][\depth]{\large\scshape Abstract}}%
    \vspace{0.5ex}\begin{quote}}{\par\end{quote}}

\begin{document}
\raggedbottom
\maketitle
\lhead{Preprint}
\begingroup
\renewcommand{\thefootnote}{\fnsymbol{footnote}}
\footnotetext[1]{Equal contribution. \quad
    $^\dagger$Corresponding author.}
\endgroup

\begin{abstract}
Video segmentation models maintain object identities by carrying instance information across
frames. Under prolonged occlusion, reappearance, or interactions between similar instances,
however, an unreliable update can overwrite a valid history and cause persistent identity drift.
We introduce \textbf{POSReasoner}, a trainable, plug-and-play framework that explicitly decides
when an observation should change an object's state. Each persistent state records identity,
confidence, and absence history. A sparse state--observation graph supports Propose--Verify
reasoning: provisional associations are revisited using object history, predicted presence, and
competition among identities. The verified decisions determine whether to retain, update,
reactivate, or suppress each state, while a learned gate controls the evidence written back to
memory. Only verified transitions update the persistent state used in subsequent frames.
POSReasoner uses standard video annotations and keeps the base model frozen, enabling integration
with diverse VOS and VIS architectures. Experiments across long-term VOS and VIS benchmarks show
consistent improvements over strong baselines, with the largest gains under occlusion and object
reappearance.
\end{abstract}

\section{Introduction}

\begin{wrapfigure}{r}{0.40\textwidth}
    \vspace{-0.7\baselineskip}
    \centering
    \includegraphics[width=\linewidth]{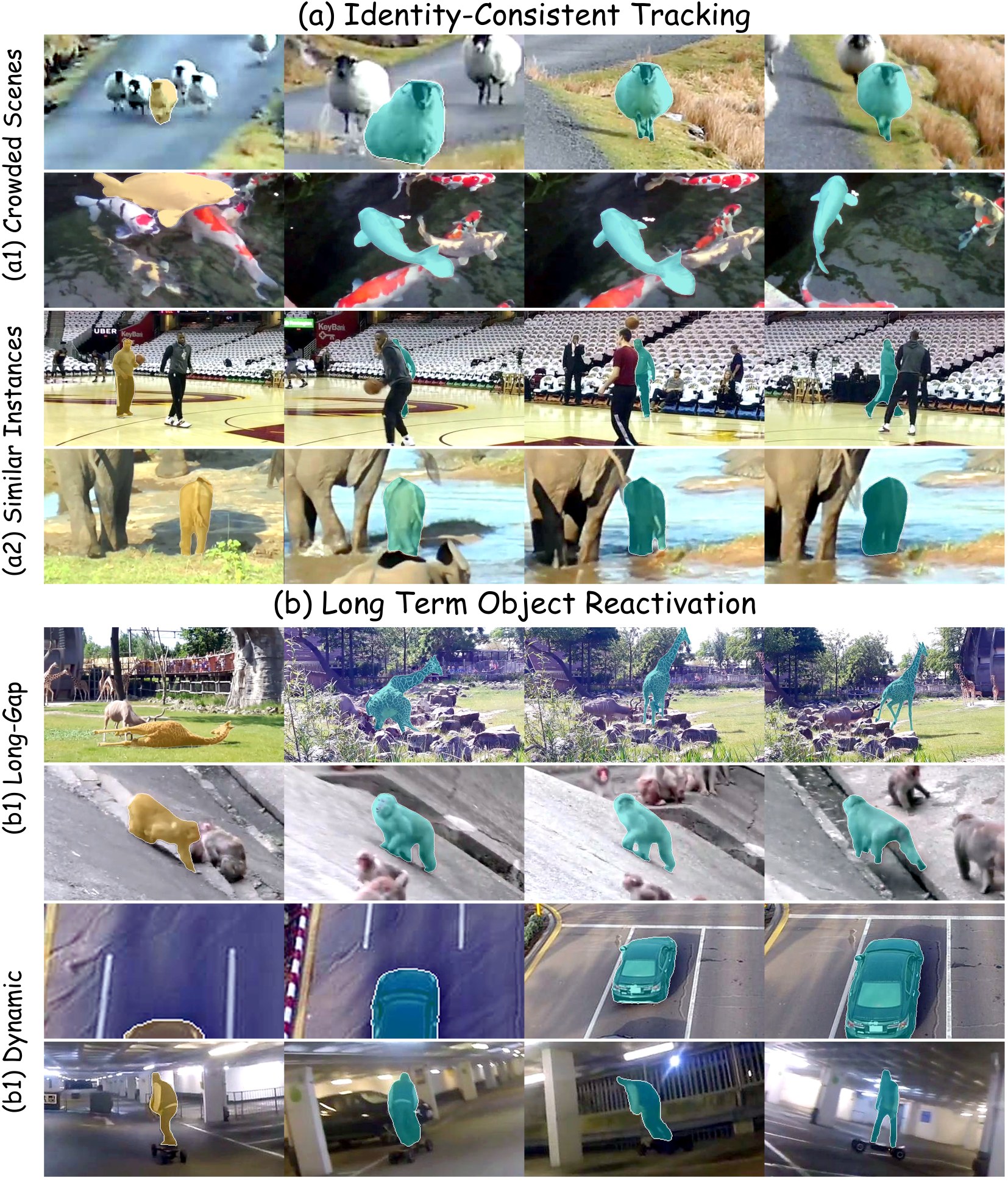}
    \captionsetup{font=footnotesize,skip=2pt}
    \caption{POSReasoner preserves identity under crowding and similar-instance
    interactions (a), and reactivates returning objects after long gaps or
    dynamic motion (b). Yellow denotes references; teal denotes predictions.}
    \label{fig:teaser}
    \vspace{-0.3\baselineskip}
\end{wrapfigure}

Video segmentation must determine not only what occupies each frame, but also whether masks
separated in time belong to the same object. This temporal identity requirement is shared by
video object segmentation (VOS), where target objects are specified, and video instance
segmentation (VIS), where they must also be discovered and classified \citep{youtubevis,ovis}.
Modern segmentation backbones and video-level object representations produce increasingly
accurate masks \citep{mask2former,video_mask2former,sam2}. Identity continuity, however, remains
brittle under prolonged occlusion, reappearance, and interactions between similar instances,
when current observations are least reliable and a mistaken association can affect the rest of
the video.

Existing methods preserve temporal continuity by carrying features or object queries across
frames \citep{wang2026stc,wen2026evostreaming,wen2025ai4service,li2026dvlt,li2026videococo}.
Memory-based VOS systems retrieve features from
prior frames \citep{xmem,cutie,sam2},
while query-based and decoupled VIS systems propagate queries, link predictions, or refine tracks
\citep{seqformer,idol,minvis,vita,genvis,dvis,dvispp}. Recent designs strengthen video-frame
interaction and contextual association \citep{syncvis,cavis}, manage object memory explicitly
\citep{lomm}. DAM4SAM filters distractors, whereas SAM2Long retains multiple segmentation
pathways \citep{videnovic2025dam4sam,sam2long}. Despite this progress, three coupled limitations
remain.
First, filtering and pathway search reduce bad evidence, but selected observations still condition
later predictions; once an error is admitted, subsequent matching uses an altered identity
reference \citep{videnovic2025dam4sam,sam2long}. Second, a missing match may indicate occlusion,
true absence, or a missed detection; suppressing low-confidence evidence alone does not determine
whether an old identity should persist or reactivate. LOMM models object presence and LTMU predicts
update readiness \citep{lomm,dai2020ltmu}, but these signals are not jointly reasoned with
multi-object association and state admission. Third, synchronized, contextual, and decoupled
designs improve candidate association \citep{dvis,dvispp,syncvis,cavis}, but do not revisit
provisional matches with lifecycle state before write-back; competing identities may therefore be
locally plausible yet mutually inconsistent. These limitations share a root cause: the current
observation is treated as the next state rather than evidence for a state transition.

To address these limitations, we introduce \textbf{POSReasoner}, a trainable, plug-and-play
framework that reasons over persistent, identity-indexed object states. To protect reliable
history, each state records identity evidence, geometry, visibility history, and confidence, and
persists by default until a transition is verified. To model absence and reappearance explicitly,
a sparse state--observation graph links active and absent states to the current candidates and a
learned null observation. Reasoning then follows a Propose--Verify procedure. \emph{Propose}
recalls the object history and forms provisional state--observation associations, while
\emph{Verify} revisits them using predicted object presence and excess candidate demand.
Gated residual corrections allow verification to resolve duplicate claims without discarding a
competent proposal. A transition head finally retains, updates, reactivates, or suppresses each
state. Association, presence, and transition estimates receive intermediate supervision, but the
internal steps do not advance video time: only the final verified transition writes candidate
evidence into the persistent state. POSReasoner therefore turns association, lifecycle inference,
and memory admission into a single recurrent
state transition rather than a sequence of disconnected decisions. The verified transition then
conditions the next frame, allowing reasoning to shape future association and segmentation rather
than merely revise the current output. All reasoning targets come from standard video annotations,
while the host model remains frozen.

By jointly leveraging persistent object states, conflict-aware verification, and selective
write-back, POSReasoner maintains reliable identities through occlusion and reappearance, as shown
in Figure~\ref{fig:teaser}. The lightweight framework augments structurally different
host architectures without redesigning their segmentation or tracking components. Experiments
demonstrate consistent improvements over strong hosts, with the clearest gains under severe
occlusion and object reappearance.

In summary, our main contributions are as follows:
\begingroup
\setlength{\leftmargini}{1.45em}
\begin{itemize}
    \item \textbf{Persistent-State Formulation.} We formulate long-horizon identity maintenance
    over persistent object states and introduce POSReasoner, a trainable framework that keeps
    VOS/VIS hosts frozen.
    \item \textbf{Propose--Verify Reasoning.} A recurrent sparse state--observation graph jointly
    infers presence, resolves competing associations, and admits only verified transitions to
    memory.
    \item \textbf{Broad Empirical Validation.} Across long-term VOS/VIS benchmarks, POSReasoner
    consistently improves strong hosts, especially under occlusion and reappearance; focused
    analyses isolate gains from persistent states and learned transitions.
\end{itemize}
\endgroup

\section{Related Work}

\subsection{Video Instance Segmentation}
Video instance segmentation (VIS) jointly predicts object categories, masks, and identities
\citep{youtubevis,ovis}. Video-level methods aggregate frame or clip queries
\citep{video_mask2former,seqformer,vita}, whereas online methods associate frame predictions through
query propagation or learned embeddings \citep{idol,minvis}. GenVIS and CTVIS strengthen temporal
association with propagated prototypes and memory banks \citep{genvis,ctvis}; DVIS and DVIS++
decouple segmentation, tracking, and temporal refinement \citep{dvis,dvispp}. Recent methods further
introduce dynamic anchors, synchronized queries, and contextual matching
\citep{dvisdaq,syncvis,cavis}. LOMM maintains a presence-aware latest-object memory and separates
existing-object association from new-object allocation \citep{lomm}. Tracking instability
nonetheless remains pronounced in long, crowded, and heavily occluded videos
\citep{mindthegap}. Together, these designs make associations more reliable, yet association and
state update are usually coupled: once an observation is matched, it becomes part of the reference
used in later frames. A local association error can therefore alter the identity evidence on which
subsequent decisions depend.

\subsection{Memory-Based Video Object Segmentation}
Memory-based VOS propagates reference-frame objects through past observations. STM and STCN
establish dense space--time correspondence \citep{stm,stcn}, while AOT and DeAOT propagate
multiple objects through identification embeddings \citep{aot,deaot}. XMem separates sensory,
working, and long-term memory \citep{xmem}, and Cutie combines pixel memory with object queries
\citep{cutie}. Foundation models retain a similar recurrent interface: SAM~2 uses streaming memory
attention \citep{sam2}, while SAM~3 combines image-level detection with a memory-based tracker and
an object-presence head \citep{sam3}. Extensions for long videos retain alternative mask pathways,
suppress distractor-contaminated memory, retrieve memories using motion and spatiotemporal cues, or
make SAM~3's memory selection object-specific
\citep{sam2long,videnovic2025dam4sam,mosam,sam3dms}. These methods improve which observations are
stored or retrieved within a particular memory design. Our focus is complementary: given the
candidates exposed by an existing segmenter, we model their competing claims on persistent
identity states and determine the resulting state revision.

\subsection{Object-Centric Video Learning}
Object-centric models decompose a video into recurrent object slots. SAVi propagates slots using
motion and initialization cues \citep{savi}. Dual-State Slot Attention separates transient
appearance from persistent identity and filters identity updates through a learned transition
\citep{dual_state_slot_attention}, while Temporal Slot Activation predicts whether a slot is active
and gates both its update and decoding during invisibility \citep{temporal_slot_activation}. Recent
open-world segmentation also combines hierarchical mask discovery, deferred object admission, and
track consolidation for long-range identity maintenance
\citep{openworld_video_segmentation}. The connection to our setting lies in their persistent
representation of identity. Their learning objectives, however, concern scene decomposition or
open-world discovery rather than the states maintained by an existing segmentation model. Our work
addresses this missing state-revision step: it starts from candidates supplied by a frozen host,
reasons over their claims on persistent object states, and carries only the verified transitions
forward.

\begingroup
\setlength{\textfloatsep}{10pt plus 2pt minus 2pt}
\section{Method}
\label{sec:method}
\begingroup
\setlength{\abovedisplayskip}{3pt plus 1pt minus 1pt}
\setlength{\belowdisplayskip}{3pt plus 1pt minus 1pt}
\setlength{\abovedisplayshortskip}{2pt plus 1pt}
\setlength{\belowdisplayshortskip}{2pt plus 1pt minus 1pt}

Figure~\ref{fig:paradigm_compare} summarizes the transition from frame-local
association and direct memory updates to our persistent-state formulation,
which verifies competing hypotheses before admitting evidence to memory.

\begin{figure}[!t]
    \centering
    \includegraphics[width=\textwidth]{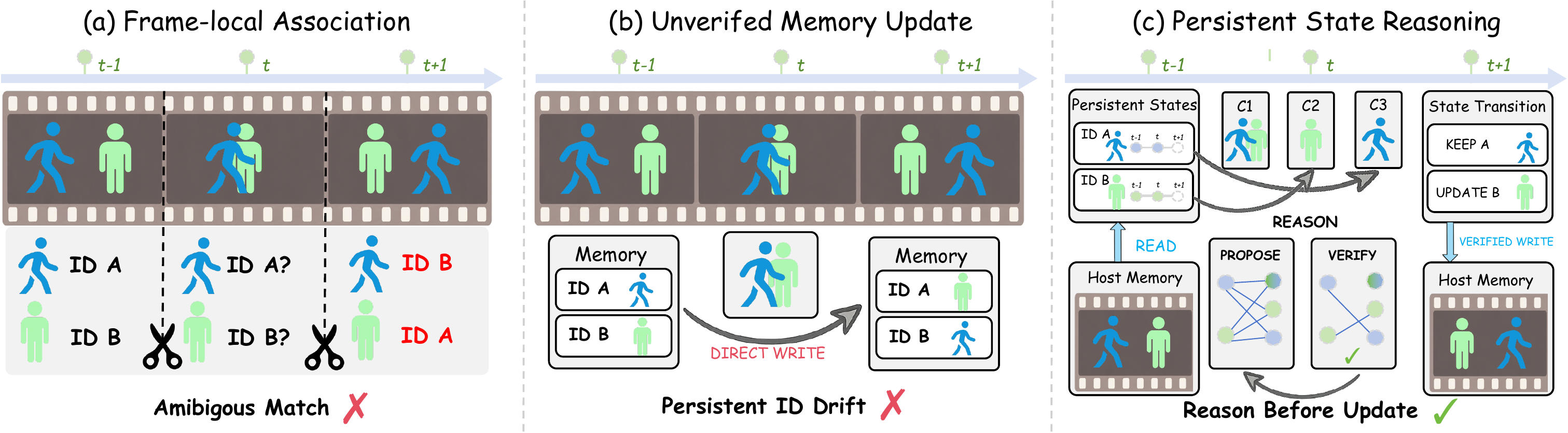}
    \caption{From association to persistent-state reasoning. (a) Frame-local
    association leaves ambiguous matches unresolved across time. (b) Direct
    write-back can propagate one unreliable observation through later states.
    (c) POSReasoner keeps identity-indexed states separate from host memory,
    verifies competing hypotheses, and applies an explicit state transition.}
    \label{fig:paradigm_compare}
\end{figure}

Given a video $\mathcal{X}=\{I_t\}_{t=1}^{T}$, a frozen host model produces a
set of object observations at every frame. They correspond to propagated
queries in VIS \citep{dvis,dvispp,lomm} or memory readouts in VOS
\citep{xmem,sam2,sam3}. A match does not by itself determine a memory update.
Depending on the object's history and competing identities, the matched
observation may update a visible state, revive an absent one, or be rejected.
POSReasoner treats each association as a \textbf{state-transition hypothesis}.
The problem is thus not only which observation matches, but whether and how
that match should alter the persistent state.

As illustrated in Figure~\ref{fig:method_overview}(a), POSReasoner pairs frozen-host
observations with persistent states that retain identity evidence and lifecycle
history. Propose forms provisional associations, while Verify uses object
competition and state history to refine both identity assignments and transition
actions (b). These decisions gate memory write-back, admitting candidate evidence
in proportion to the support for an update or reactivation (c). Intermediate
hypotheses remain in a temporary workspace: only the final verified transition
updates the persistent state carried to the next frame. Thus, a proposal can be
revised without prematurely altering persistent memory.

\subsection{Persistent Object States}
\label{sec:persistent_states}

For frame $I_t$, the host $\mathcal{H}$ returns object features,
class probabilities, and masks as
$[\widetilde{\mathbf{Q}}_t,\mathbf{P}_t,\mathbf{M}_t]=\mathcal{H}(I_t)$.
The adapter represents row $j$ by a host feature $q_t^j$ and appends cues
already available from the host: foreground confidence, class confidence,
class margin, and predicted mask quality. VIS queries and VOS memory readouts
thus share the same state--observation interface.

Before processing $I_t$, the state bank
$\mathcal{S}_{t-1}=\{s_{t-1}^i\}_{i=1}^{N_{t-1}}$ stores one entry for every
admitted identity. Each entry keeps a latent identity feature together with
state confidence and time since the last reliable observation. VOS states are
initialized at the annotated first appearance. In
VIS, a foreground observation that is not assigned to an existing identity
opens a free state slot. During absence, the lifecycle record advances while
the identity feature is left intact. State confidence is updated only for a
visible observation; otherwise its previous value is retained and the absence
duration is incremented.

Let $\eta_i$ collect the state metadata and $\mu_j$ the observation metadata.
We suppress time indices below, writing $z_i=z_{t-1}^i$ and $q_j=q_t^j$.
Two lightweight projections map them to the common reasoning dimension:
$h_i^0=f_s([z_i;\eta_i])$ and $u_j^0=f_o([q_j;\mu_j])$.
Here $z_i$ is the persistent identity feature and $q_j$ is the current
frozen-host feature. The projections are trained with POSReasoner, while the
host features remain frozen. In our implementation, $\eta_i$ contains state
confidence and normalized absence duration, and $\mu_j$ contains visual
confidence, class confidence, class margin, and mask quality.

We connect the state and observation sets with a sparse bipartite graph. Its
edges retain globally confident observations and the top-$K$
identity-compatible observations for each state. A learned null observation is
connected to every state, so occlusion need not be explained by a forced match.
Every real edge is treated as a candidate association whose effect on the
persistent state is determined jointly with the transition action; the null
edge represents leaving the state unmatched.

\begin{figure}[!t]
    \centering
    \includegraphics[width=\textwidth]{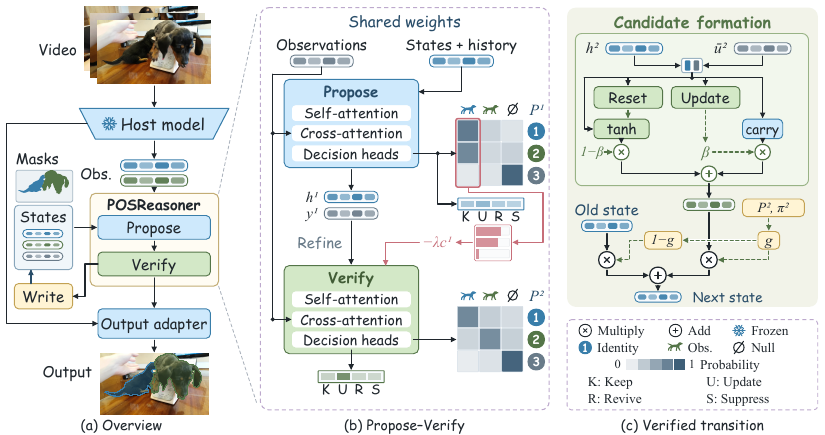}
    \caption{Overview of POSReasoner. (a) Frozen-host observations are paired
    with persistent object states. (b) Propose--Verify reasoning refines
    associations and actions using identity history and object competition.
    The shaded matrices associate identities with observations or null.
    (c) Verified decisions gate a single state update.
    Internal values are schematic; \href{https://davischallenge.org/}{DAVIS}
    frames with ground-truth masks illustrate the interface, not model predictions.}
    \label{fig:method_overview}
\end{figure}

\vspace{-0.3em}
\subsection{Propose--Verify State Reasoning}
\label{sec:state_observation_reasoning}
\vspace{-0.3em}

A locally plausible association may be inconsistent with the object set, as
when two identities claim the same observation. POSReasoner therefore runs two
association steps with shared parameters. At step $r$, self-attention
contextualizes the states and cross-attention reads adjacent observations
\citep{vaswani2017attention}. Let $h_i^r$ and $u_j$ be the resulting features,
and let $z_i$ and $q_j$ denote the persistent identity feature and current
frozen-host feature, respectively. Before the first step, observations are
also contextualized by self-attention. A learned
step embedding distinguishes Propose from Verify while the attention and
prediction parameters remain shared. Invalid graph edges are masked throughout
cross-attention and association. Writing
$\widetilde{\mathcal{N}}_i=\mathcal{N}_i\cup\{\varnothing\}$, we compute
\begin{align}
    e_{ij}^{r}
    &=
    \frac{\operatorname{cos}(W_s h_i^r,W_o u_j)}{\tau}
    +\lambda_{\mathrm{id}}\operatorname{cos}(z_i,q_j),\nonumber\\
    p_{ij}^{r}
    &=\operatorname*{softmax}_{j\in\widetilde{\mathcal{N}}_i}
    \left(e_{ij}^{r}-\lambda_{\mathrm{cmp}}c_j^{r-1}\right),
    \qquad
    c_j^{r}
    =\left[\sum_i p_{ij}^{r}-1\right]_+ .
    \label{eq:association}
\end{align}
Here $\mathcal{N}_i$ is the sparse neighborhood of state $i$, and
$\varnothing$ denotes the learned null observation, whose identity-prior
term is explicitly defined as zero.

At the Propose step, we set $c_j^0=0$ in
Equation~\ref{eq:association}. The score $e_{ij}^1$
combines compatibility in the contextualized feature space with the original host
identity similarity. Normalization over
$\widetilde{\mathcal{N}}_i$ gives a distribution over the retained candidates
and the null observation. The column sum $\sum_i p_{ij}^1$ then measures how
much total probability the state set assigns to observation $j$, and $c_j^1$
retains only the probability mass above one. It therefore acts as a soft
duplicate-demand penalty rather than enforcing a hard one-to-one assignment.

For state $i$, the matched evidence is
$\bar u_i^r=\sum_j p_{ij}^r u_j$. We compare it with the reasoned state using
their concatenation, element-wise product, and absolute difference. Association
entropy and probability-weighted candidate occupancy are appended to this
comparison. The decision feature is
\begin{align}
    r_i^r
    &=\big[
    h_i^r;\bar u_i^r;
    h_i^r\odot\bar u_i^r;
    |h_i^r-\bar u_i^r|;
    \mathcal{E}(p_i^r);
    \rho_i^r
    \big],\nonumber\\
    \rho_i^r
    &=\sum_j p_{ij}^r\sum_k p_{kj}^r ,
    \label{eq:decision_feature}
\end{align}
where $\mathcal{E}$ is association entropy and $\rho_i^r$ measures the demand
on the candidates preferred by state $i$. Three prediction heads applied to
$r_i^r$ estimate visibility and existence, a transition distribution $\pi_i^r$, and
\textbf{transition advantage} $v_i^r$. Together with the association logits,
these predictions form a decision tuple
$\mathbf{y}_i^r=[\boldsymbol{\ell}_i^r;\mathbf{b}_i^r;
\boldsymbol{a}_i^r;v_i^r]$,
where $\mathbf{b}_i^r$ contains visibility and existence beliefs, and
$\boldsymbol{a}_i^r$ contains the four transition logits, with
$\pi_i^r=\operatorname{softmax}(\boldsymbol{a}_i^r)$;
$\boldsymbol{\ell}_i^r=[e_{ij}^r]_j$ contains the association logits. The tuple records a
candidate identity, its current presence, the proposed state change, and its
predicted benefit relative to retaining the state. We supervise this advantage
with the corresponding quality difference,
\begin{equation}
    \Delta_i =
    Q(\hat z_t^i,\mathcal{Y}_{t:t+H}^i)
    -Q(z_{t-1}^i,\mathcal{Y}_{t:t+H}^i),
    \qquad v_i^r\approx\Delta_i ,
    \label{eq:transition_advantage}
\end{equation}
where $Q$ is the task quality over a short annotated continuation
$\mathcal{Y}_{t:t+H}^i$ when the object is decoded from the candidate or
retained state. At inference, the reasoner predicts $v_i^r$ directly from the
current hypothesis and persistent state.

At the Verify step, the association is repeated with $c_j^1$. Observation
competition penalizes contested real observations relative to null and allows their
competing states to move to alternative candidates or the null observation.
We keep $c_\varnothing^r=0$, because multiple absent states may select null
simultaneously. State confidence and absence duration remain part of $h_i^r$, so
Verify can also change the action without changing the selected observation:
the same match may imply \textsc{Update}, \textsc{Revive}, or \textsc{Keep}
for different state histories. Verify can therefore revise either the proposed
ownership or the proposed state transition. This coupled revision distinguishes
state-transition reasoning from independently reranking a fixed set of
predictions.

The second-step tuple is obtained by gated interpolation,
$\mathbf{y}_i^2=\mathbf{y}_i^1+\boldsymbol{\alpha}\odot
(\widetilde{\mathbf{y}}_i^2-\mathbf{y}_i^1)$,
where $\widetilde{\mathbf{y}}_i^2$ is predicted from the updated matched
evidence, association entropy, and observation competition. Learned coefficients
separately refine association, action, and the belief/advantage residuals. They
are initialized near zero, so Verify begins from the proposed tuple and learns
how strongly each output group should change. Both reasoning steps receive
intermediate supervision.

\subsection{Verified State Transition and Learning}
\label{sec:verified_transition}

The verified action probabilities determine the state transition. \textsc{Keep} retains a
reliable state, \textsc{Update} incorporates a visible observation,
\textsc{Revive} reconnects a previously absent object, and \textsc{Suppress}
rejects inconsistent evidence. A gated recurrent unit \citep{cho2014learning}
first forms a candidate state
$\hat z_t^i=\operatorname{GRU}(h_i^2,\bar u_i^2)$.
In Figure~\ref{fig:method_overview}(c), Reset gates the features entering
the nonlinear branch, while Update produces the carry weight $\beta$ that
blends the carry and nonlinear branches. These internal gates form the
candidate; a separate write gate controls its admission to persistent memory:
\begin{align}
    g_i
    &=
    \big(\pi_{i,\mathrm{update}}^2+\pi_{i,\mathrm{revive}}^2\big)
    \max_{j\neq\varnothing}p_{ij}^2
    \big(1-p_{i\varnothing}^2\big),\nonumber\\
    z_t^i
    &=(1-g_i)z_{t-1}^i+g_i\hat z_t^i .
    \label{eq:state_transition}
\end{align}
The gate combines the probability of a writing transition, the strongest real
association, and the total non-null mass. Otherwise, the persistent
state is carried forward. Probability assigned to \textsc{Keep} or
\textsc{Suppress} does not contribute to $g_i$; as either action dominates,
the state write approaches zero. Both reasoning steps operate on a temporary
workspace anchored to $z_{t-1}^i$; intermediate proposals are never committed
to temporal memory. This prevents the same frame from updating an identity
multiple times. Only the final transition advances the state at frame $t$.
Appendix~\ref{sec:mathematical_properties} derives the properties of association,
verification, and memory admission.

At inference, conflicting non-null claims are ranked using association
confidence and transition advantage. The advantage estimates whether a proposed
write is preferable to retaining the current state; it does not alter the sparse
candidate graph. For assignment-based hosts, the output adapter retains only
the highest-ranked claim to each real observation. The verified state and its
visibility record are carried to
$t+1$ and become the reference for the next association.

The verified decision is also returned through a host-specific output adapter.
Residual hosts blend a query toward the proposed or matched state according to
the \textsc{Update} and \textsc{Revive} probabilities, while assignment-based
hosts transfer the selected observation to its state slot. After existing states
have been assigned, valid foreground observations that remain unclaimed are
allocated to free slots. The null observation never consumes a slot and may be
selected by multiple absent states.

Training targets are constructed from frozen-host predictions and annotations
on the training split. The association target is the observation with the same
identity, or the null observation when the object is absent. A visible match is
labeled \textsc{Update}, and a match after absence is labeled \textsc{Revive}.
An observed state without a valid identity receives \textsc{Suppress}; an
unobserved state without a reliable candidate receives \textsc{Keep}. The
advantage target is the signed quality difference in
Equation~\ref{eq:transition_advantage}.
\begin{align}
    \mathcal{L}_{\mathrm{POSR}}
    &=
    \sum_{r=1}^{2}\omega_r\big(
    \mathcal{L}_{\mathrm{assoc}}^r
    +\lambda_{\mathrm{tr}}\mathcal{L}_{\mathrm{tr}}^r
    +\lambda_{\mathrm{bel}}\mathcal{L}_{\mathrm{bel}}^r
    \big)\nonumber\\
    &+\lambda_{\mathrm{state}}\mathcal{L}_{\mathrm{state}}
    +\lambda_{\mathrm{adv}}\mathcal{L}_{\mathrm{adv}}
    +\lambda_{\mathrm{ref}}\mathcal{L}_{\mathrm{ref}} .
    \label{eq:training_objective}
\end{align}
$\mathcal{L}_{\mathrm{assoc}}$ and $\mathcal{L}_{\mathrm{tr}}$ are
cross-entropy losses, while $\mathcal{L}_{\mathrm{bel}}$ sums binary
cross-entropy terms for visibility and existence. $\mathcal{L}_{\mathrm{state}}$
is the cosine distance to a
stop-gradient target: the matched observation embedding for a writing
transition and the retained identity feature otherwise.
$\mathcal{L}_{\mathrm{adv}}$ is a smooth-$L_1$ loss on transition advantage,
and the refinement loss is
$\mathcal{L}_{\mathrm{ref}}=|\mathcal{V}|^{-1}\sum_{i\in\mathcal{V}}
[p_{i,y_i}^{1}-p_{i,y_i}^{2}]_+$,
where $y_i$ is the target association and $\mathcal{V}$ is the valid-state set.
The verified step receives the larger weight. At validation and test time,
POSReasoner uses only host outputs, its persistent state, and task-provided
prompts.
\endgroup

\endgroup
\section{Experiments}
\label{sec:experiments}
\setcounter{topnumber}{1}
\setlength{\textfloatsep}{8pt plus 2pt minus 2pt}

\subsection{Experimental Settings}

We evaluate VIS on YouTube-VIS 2019, 2021, and 2022~\citep{youtubevis}
and on OVIS~\citep{ovis}, reporting mask AP, AP50, and AP75. YouTube-VIS
covers diverse video lengths and category distributions, while OVIS emphasizes
crowded scenes and inter-object occlusion. For long-term VOS, we use LVOS v1
and v2~\citep{hong2023lvos,hong2024lvosv2}, with region similarity
$\mathcal{J}$, contour accuracy $\mathcal{F}$, and their mean
$\mathcal{J}\&\mathcal{F}$~\citep{perazzi2016davis}. These benchmarks evaluate
identity maintenance both with and without object discovery and category
prediction.

For VIS, we integrate POSReasoner with CTVIS~\citep{ctvis},
DVIS++~\citep{dvispp}, DVIS-DAQ~\citep{dvisdaq}, and LOMM~\citep{lomm},
representing different association and memory designs. GenVIS~\citep{genvis}
and other published methods provide additional comparisons. For VOS, we use
SAM3~\citep{sam3}. Paired comparisons keep the host checkpoint, candidate
masks, input resolution, and official evaluator fixed. Only POSReasoner is
trained, using training annotations; the host remains frozen.

Training uses video-grouped folds to keep each video within one split.
Out-of-fold predictions select the readout before model ensembling.
For hosts without frame-wise identity sequences, we use the available state
and relation descriptors under the same protocol. The VOS readout uses
disjoint fitting, calibration, and holdout videos. All architecture,
optimization, and candidate-budget choices are fixed before evaluation.

\subsection{Comparison with State-of-the-Art Methods}
\label{sec:vis_results}

\begin{figure}[!t]
    \centering
    \includegraphics[width=\textwidth]{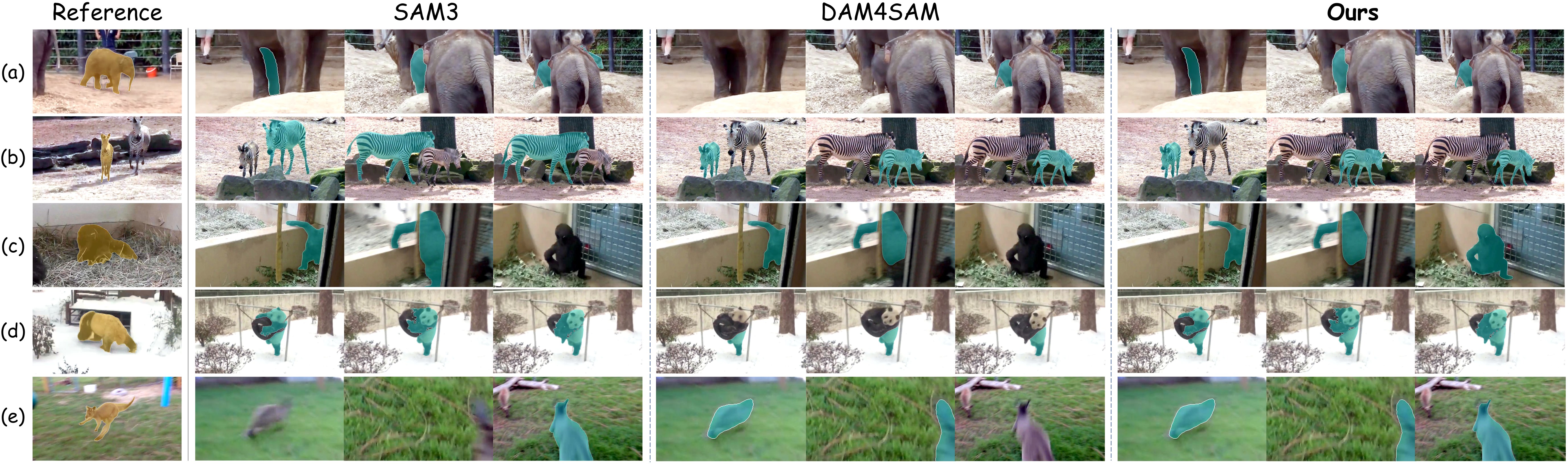}
    \captionsetup{
        font=small,
        width=\textwidth,
        skip=3pt,
        justification=centering
    }
    \caption{Qualitative comparison under occlusion, identity ambiguity, and
    object reappearance.}
    \label{fig:qualitative}
\end{figure}

\noindent\textbf{Results on YouTube-VIS.}
POSReasoner consistently improves AP across the three benchmarks
(Table~\ref{tab:ytvis_main_preview}). With ResNet-50, CTVIS gains 1.0,
1.1, and 2.1 AP on YouTube-VIS 2019, 2021, and 2022, respectively;
DVIS++ and DAQ also improve on each benchmark. With ViT-L, LOMM gains
0.7, 0.8, and 1.0 AP. These paired improvements, obtained with fixed
host weights and candidate masks, show that state reasoning complements
different association and memory designs. Figure~\ref{fig:qualitative}
illustrates identity preservation under occlusion, similar-object
interactions, and reappearance.

\begin{table}[!htbp]
    \centering
    \captionsetup{
        font=small,
        width=0.96\textwidth,
        skip=3pt,
        justification=raggedright,
        singlelinecheck=false
    }
    \caption{YouTube-VIS validation. Published methods provide context; each
    highlighted row adds POSReasoner to the host immediately above. Bold and
    underline mark the best and second-best result within each backbone.}
    \label{tab:ytvis_main_preview}
    \begingroup
    \scriptsize
    \setlength{\tabcolsep}{2.2pt}
    \renewcommand{\arraystretch}{0.92}
    \ifdefined\vismaintablestyle\vismaintablestyle\fi
    \begin{tabularx}{0.96\textwidth}{@{}lX*{9}{c}@{}}
        \toprule
        \rowcolor{black!5}
        \multirow{2}{*}{\textbf{Backbone}} &
        \multirow{2}{*}{\textbf{Method}} &
        \multicolumn{3}{c}{\textbf{YTVIS 2019}} &
        \multicolumn{3}{c}{\textbf{YTVIS 2021}} &
        \multicolumn{3}{c}{\textbf{YTVIS 2022}} \\
        \cmidrule(lr){3-5}
        \cmidrule(lr){6-8}
        \cmidrule(lr){9-11}
        \rowcolor{black!5}
        & & AP & AP50 & AP75 & AP & AP50 & AP75 & AP & AP50 & AP75 \\
        \midrule
        \multirow{7}{*}{R50}
        & GenVIS~\citep{genvis}
        & 50.0 & 71.5 & 54.6 & 47.1 & 67.5 & 51.5 & 37.5 & 61.6 & 41.5 \\
        & CTVIS~\citep{ctvis}
        & 51.8 & 74.1 & 55.2 & 49.5 & 72.5 & 53.6 & 45.4 & 67.5 & 48.9 \\
        \rowcolor{POSRbg}
        & \hspace{0.4em}\textcolor{POSRaccent}{\textbf{+ POSReasoner}}
        & 52.8 & 75.2 & 56.5
        & 50.6 & \textbf{73.7} & 55.1
        & \textbf{47.5} & \textbf{69.7} & \textbf{51.4} \\
        & DVIS++~\citep{dvispp}
        & \underline{55.7} & \textbf{80.8} & 59.8
        & 50.4 & 71.4 & 55.2
        & 46.3 & 66.6 & 50.5 \\
        \rowcolor{POSRbg}
        & \hspace{0.4em}\textcolor{POSRaccent}{\textbf{+ POSReasoner}}
        & \textbf{56.0} & \underline{80.6} & \textbf{60.9}
        & \textbf{51.1} & 72.6 & \underline{55.8}
        & \underline{47.0} & \underline{68.2} & \underline{51.0} \\
        & DAQ~\citep{dvisdaq}
        & 54.6 & 78.2 & \underline{60.3}
        & 50.1 & 71.6 & 55.1
        & 45.9 & 66.7 & 50.0 \\
        \rowcolor{POSRbg}
        & \hspace{0.4em}\textcolor{POSRaccent}{\textbf{+ POSReasoner}}
        & 55.5 & 79.2 & \textbf{60.9}
        & \underline{51.0} & \underline{72.9} & \textbf{55.9}
        & 46.7 & 67.9 & 50.9 \\
        \midrule
        \multirow{3}{*}{ViT-L}
        & DVIS++~\citep{dvispp}
        & 67.7 & 88.8 & 75.3 & 62.3 & 82.7 & 70.2 & 37.5 & 53.7 & 39.4 \\
        & LOMM~\citep{lomm}
        & \underline{69.1} & \underline{89.3} & \underline{76.5}
        & \underline{65.0} & \underline{85.8} & \underline{72.7}
        & \underline{60.4} & \underline{82.7} & \underline{67.1} \\
        \rowcolor{POSRbg}
        & \hspace{0.4em}\textcolor{POSRaccent}{\textbf{+ POSReasoner}}
        & \textbf{69.8} & \textbf{89.5} & \textbf{77.0}
        & \textbf{65.8} & \textbf{86.0} & \textbf{73.3}
        & \textbf{61.4} & \textbf{83.4} & \textbf{67.6} \\
        \bottomrule
    \end{tabularx}
    \endgroup
\end{table}

\noindent\textbf{Results on OVIS.}
The gains are larger on occlusion-heavy OVIS
(Table~\ref{tab:ovis_main_preview}). With ResNet-50, CTVIS, DVIS++, and
DAQ improve by 2.7, 2.1, and 1.6 AP, respectively. Each host gains more
than on any YouTube-VIS benchmark, consistent with the value of retaining
identity history when current observations are ambiguous. Improvements of
3.2, 1.9, and 1.8 AP75 also show that the benefits extend to a stricter
mask-overlap threshold.

With larger backbones, DAQ improves from 49.6 to 50.6 AP with Swin-L
and from 53.9 to 55.2 AP with ViT-L, with gains in both AP50 and AP75.
State reasoning thus remains beneficial alongside stronger visual
representations in crowded, occluded scenes.

\begingroup
\setlength{\intextsep}{3pt}
\begin{table}[H]
    \centering
    \captionsetup{
        font=small,
        width=0.94\textwidth,
        skip=3pt,
        justification=raggedright,
        singlelinecheck=false
    }
    \caption{OVIS validation. Highlighted rows add POSReasoner to the matched
    host; $\Delta$AP is relative to that host. Bold and underline mark the best
    and second-best results within each backbone.}
    \label{tab:ovis_main_preview}
    \begingroup
    \small
    \setlength{\tabcolsep}{5pt}
    \renewcommand{\arraystretch}{0.94}
    \begin{tabularx}{0.94\textwidth}{
        @{}X
        C
        *{3}{S[table-format=2.1]}@{}
    }
    \toprule
    \textbf{Method} &
    \textbf{$\Delta$AP} &
    \multicolumn{1}{c}{\textbf{AP}} &
    \multicolumn{1}{c}{\textbf{AP50}} &
    \multicolumn{1}{c}{\textbf{AP75}} \\
    \midrule

    \rowcolor{black!5}
    \multicolumn{5}{@{}l}{\strut\textbf{ResNet-50}} \\
    CTVIS~\citep{ctvis} & -- & 34.6 & 59.9 & 33.8 \\
    \rowcolor{POSRbg}
    \hspace{1em}\textcolor{POSRaccent}{\textbf{+ POSReasoner}}
        & \gain{+2.7} & 37.3 & 63.2 & 37.0 \\
    \addlinespace[1pt]
    DVIS++~\citep{dvispp} & -- & 36.8 & 61.3 & 37.5 \\
    \rowcolor{POSRbg}
    \hspace{1em}\textcolor{POSRaccent}{\textbf{+ POSReasoner}}
        & \gain{+2.1} & \second{38.9} & 64.3 & \best{39.4} \\
    \addlinespace[1pt]
    DAQ~\citep{dvisdaq} & -- & 38.3 & \second{65.2} & 37.3 \\
    \rowcolor{POSRbg}
    \hspace{1em}\textcolor{POSRaccent}{\textbf{+ POSReasoner}}
        & \gain{+1.6} & \best{39.9} & \best{66.7} & \second{39.1} \\

    \addlinespace[2pt]
    \rowcolor{black!5}
    \multicolumn{5}{@{}l}{\strut\textbf{Swin-L}} \\
    GenVIS~\citep{genvis} & -- & 45.2 & 69.1 & 48.4 \\
    LOMM~\citep{lomm} & -- & 47.8 & 73.6 & 51.4 \\
    DAQ~\citep{dvisdaq} & --
        & \second{49.6} & \second{76.0} & \second{52.8} \\
    \rowcolor{POSRbg}
    \hspace{1em}\textcolor{POSRaccent}{\textbf{+ POSReasoner}}
        & \gain{+1.0} & \best{50.6} & \best{77.0} & \best{53.9} \\

    \addlinespace[2pt]
    \rowcolor{black!5}
    \multicolumn{5}{@{}l}{\strut\textbf{ViT-L}} \\
    DVIS++~\citep{dvispp} & -- & 49.6 & 72.5 & 55.0 \\
    LOMM~\citep{lomm} & -- & 51.7 & 73.9 & 57.5 \\
    DAQ~\citep{dvisdaq} & --
        & \second{53.9} & \second{78.9} & \second{58.5} \\
    \rowcolor{POSRbg}
    \hspace{1em}\textcolor{POSRaccent}{\textbf{+ POSReasoner}}
        & \gain{+1.3} & \best{55.2} & \best{79.7} & \best{59.9} \\
    \bottomrule
    \end{tabularx}
    \endgroup
\end{table}
\endgroup

\subsection{Long-Horizon State Analysis}
\label{sec:ablation}

We study how state reasoning supports identity maintenance over long videos.
Starting from the frozen SAM3 host, we progressively add state reasoning,
reactivation, and cross-trajectory reasoning on LVOS v1, and evaluate the
complete model on LVOS v2 (Table~\ref{tab:vos_main}).

\begin{table}[!t]
    \centering
    \captionsetup{
        font=footnotesize,
        width=0.86\textwidth,
        skip=3pt,
        justification=centering
    }
    \caption{Long-term VOS and cumulative component analysis with frozen SAM3
    predictions.}
    \label{tab:vos_main}
    \begingroup
    \footnotesize
    \setlength{\tabcolsep}{3.6pt}
    \renewcommand{\arraystretch}{0.86}
    \begin{tabularx}{0.86\textwidth}{@{}Xcccccc@{}}
        \toprule
        \rowcolor{black!5}
        Variant & State & React. & Cross
        & $\mathcal{J}$ & $\mathcal{F}$ & $\mathcal{J}\&\mathcal{F}$ \\
        \midrule
        \rowcolor{black!5}
        \multicolumn{7}{@{}l}{\strut\textbf{LVOS v1 cumulative components}} \\
        SAM3 host~\citep{sam3}
        & {\ding{55}} & {\ding{55}} & {\ding{55}}
        & 79.5 & 90.6 & 85.1 \\
        + state reasoning
        & {$\checkmark$} & {\ding{55}} & {\ding{55}}
        & 82.1 & 92.4 & 87.2 \\
        + reactivation
        & {$\checkmark$} & {$\checkmark$} & {\ding{55}}
        & \underline{83.3} & \underline{93.6} & \underline{88.4} \\
        \rowcolor{POSRbg}
        \textcolor{POSRaccent}{\textbf{Full POSReasoner}}
        & {$\checkmark$} & {$\checkmark$} & {$\checkmark$}
        & \textbf{83.4} & \textbf{93.7} & \textbf{88.6} \\
        \addlinespace[2pt]
        \rowcolor{black!5}
        \multicolumn{7}{@{}l}{\strut\textbf{LVOS v2 transfer}} \\
        SAM3 host~\citep{sam3}
        & {\ding{55}} & {\ding{55}} & {\ding{55}}
        & 83.4 & 91.0 & 87.2 \\
        \rowcolor{POSRbg}
        \textcolor{POSRaccent}{\textbf{Full POSReasoner}}
        & {$\checkmark$} & {$\checkmark$} & {$\checkmark$}
        & \textbf{85.3} & \textbf{92.7} & \textbf{89.0} \\
        \bottomrule
    \end{tabularx}
    \endgroup
\end{table}

\noindent\textbf{Persistent states and reactivation.}
State reasoning improves $\mathcal{J}\&\mathcal{F}$ by 2.1 points.
Its central role is to separate an object's identity history from its
current visibility: the identity feature remains available during absence,
providing a reference when observations become reliable again. Building on
these states, reactivation yields a further 1.2-point improvement. This
progression highlights two complementary requirements of long-term
segmentation: preserving an identity through absence and reconnecting the
returning object to that identity. The former retains the information
needed for later association; the latter uses that information to resume
the object's trajectory. Thus, the cumulative improvement supports treating
absence and return as explicit state transitions.

\noindent\textbf{Cross-trajectory reasoning.}
Adding cross-trajectory reasoning gives the best results across all three
metrics, bringing the full model 3.5 points above SAM3 in
$\mathcal{J}\&\mathcal{F}$. This component extends the decision from an
individual state--observation match to competition among identities.
When several states favor the same observation, verification can
redistribute their support before the persistent states are updated.
It therefore complements state retention and reactivation by considering
whether an association is consistent with the other objects in the scene.
Together, these components connect identity history, lifecycle transitions,
and object competition within the same reasoning process.

\begin{figure}[H]
    \centering
    \includegraphics[width=0.88\textwidth]{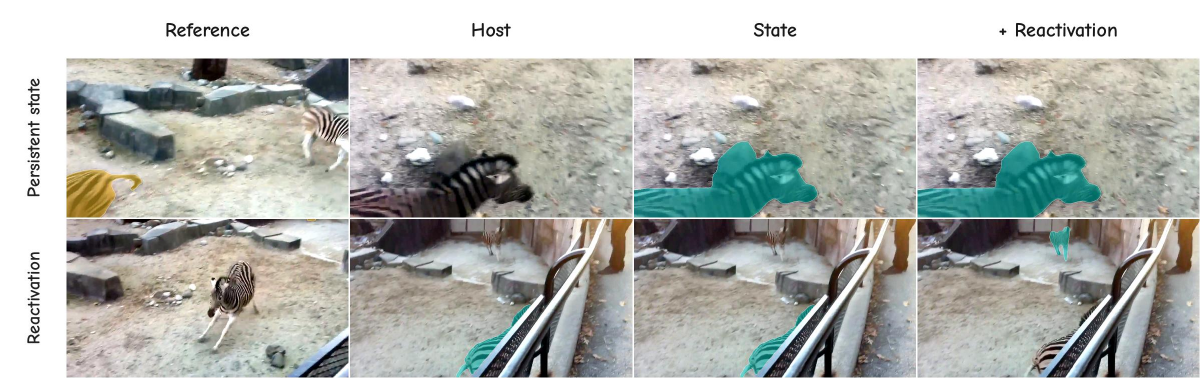}
    \captionsetup{
        font=footnotesize,
        width=0.86\textwidth,
        skip=2pt,
        justification=centering
    }
    \caption{LVOS v1: persistent state preserves identity through absence
    (top); reactivation reconnects it on return (bottom).}
    \label{fig:lvos_ablation}
\end{figure}

\noindent\textbf{Generalization and qualitative analysis.}
On LVOS v2, the complete model improves $\mathcal{J}\&\mathcal{F}$ from
87.2 to 89.0, with gains in both region similarity and contour accuracy.
The improvements across both benchmarks support the effectiveness of the
combined state-reasoning design. Figure~\ref{fig:lvos_ablation} further
illustrates the distinct roles of state retention and reactivation.
In the upper example, the state-based variant recovers the target missed
by the host. In the lower example, reactivation moves the prediction from
the foreground railing back to the returning target. These examples
illustrate how stored identity evidence supports object recovery when
observations become ambiguous or an object reappears.

\subsection{Further Analysis}

The largest observed post-reappearance gains occur after gaps of 11--30
frames: 13.1 and 4.5 points in $\mathcal{J}$ on LVOS v1 and v2,
respectively (Table~\ref{tab:reappearance_appendix}). This pattern is
consistent with retaining an identity reference through occlusion to
support association when the object returns. The OVIS controls further
highlight the role of state history: the full model reaches 37.31 AP,
compared with 34.67 for shuffled
states and 34.82 for host cues alone (Table~\ref{tab:vis_reasoning_appendix}).
The contrast supports using identity-specific histories beyond current
host evidence.

These benefits come with a compact reasoner: the five-member VIS ensemble
has 0.90M parameters and takes 5.64--5.80 ms
per inference on an A100 for 10--40 candidates (Table~\ref{tab:efficiency}).
The similar timings across these budgets indicate that increasing the
candidate count has little effect on the measured reasoner latency.

\section{Conclusion}

We present POSReasoner, a trainable, plug-and-play framework for maintaining
object identities in video segmentation. The framework represents each
object with a persistent state that preserves its identity history through
occlusion and absence. A Propose--Verify procedure jointly reasons about
candidate associations, object presence, and state transitions, allowing
reliable observations to update the state and returning objects to recover
their identities. With a shared state--observation interface, POSReasoner
can be integrated into different VOS and VIS architectures while keeping
the host models frozen. Experiments on YouTube-VIS, OVIS, and LVOS
demonstrate consistent improvements across the evaluated hosts, with
particularly strong gains under occlusion. Ablation studies and
qualitative comparisons highlight how state persistence and reactivation
contribute to long-term identity maintenance.

\clearpage
\bibliography{references}

@inproceedings{youtubevis,
  title={Video Instance Segmentation},
  author={Yang, Linjie and Fan, Yuchen and Xu, Ning},
  booktitle={Proceedings of the IEEE/CVF International Conference on Computer Vision (ICCV)},
  pages={5188--5197},
  year={2019}
}

@article{ovis,
  title={Occluded Video Instance Segmentation: A Benchmark},
  author={Qi, Jiyang and Gao, Yan and Hu, Yao and Wang, Xinggang and Liu, Xiaoyu and Bai, Xiang and Belongie, Serge and Yuille, Alan and Torr, Philip H. S. and Bai, Song},
  journal={International Journal of Computer Vision},
  year={2022}
}

@inproceedings{mask2former,
  title={Masked-Attention Mask Transformer for Universal Image Segmentation},
  author={Cheng, Bowen and Misra, Ishan and Schwing, Alexander G. and Kirillov, Alexander and Girdhar, Rohit},
  booktitle={Proceedings of the IEEE/CVF Conference on Computer Vision and Pattern Recognition (CVPR)},
  pages={1290--1299},
  year={2022}
}

@article{video_mask2former,
  title={{Mask2Former} for Video Instance Segmentation},
  author={Cheng, Bowen and Choudhuri, Anwesa and Misra, Ishan and Kirillov, Alexander and Girdhar, Rohit and Schwing, Alexander G.},
  journal={arXiv preprint arXiv:2112.10764},
  year={2021}
}

@inproceedings{seqformer,
  title={SeqFormer: Sequential Transformer for Video Instance Segmentation},
  author={Wu, Junfeng and Jiang, Yi and Bai, Song and Zhang, Wenqing and Bai, Xiang},
  booktitle={Proceedings of the European Conference on Computer Vision (ECCV)},
  pages={553--569},
  year={2022}
}

@inproceedings{idol,
  title={In Defense of Online Models for Video Instance Segmentation},
  author={Wu, Junfeng and Liu, Qihao and Jiang, Yi and Bai, Song and Yuille, Alan and Bai, Xiang},
  booktitle={Proceedings of the European Conference on Computer Vision (ECCV)},
  pages={588--605},
  year={2022}
}

@inproceedings{minvis,
  title={{MinVIS}: A Minimal Video Instance Segmentation Framework without Video-Based Training},
  author={Huang, De-An and Yu, Zhiding and Anandkumar, Anima},
  booktitle={Advances in Neural Information Processing Systems (NeurIPS)},
  year={2022}
}

@inproceedings{vita,
  title={{VITA}: Video Instance Segmentation via Object Token Association},
  author={Heo, Miran and Hwang, Sukjun and Oh, Seoung Wug and Lee, Joon-Young and Kim, Seon Joo},
  booktitle={Advances in Neural Information Processing Systems (NeurIPS)},
  year={2022}
}

@inproceedings{genvis,
  title={A Generalized Framework for Video Instance Segmentation},
  author={Heo, Miran and Hwang, Sukjun and Hyun, Jeongseok and Kim, Hanjung and Oh, Seoung Wug and Lee, Joon-Young and Kim, Seon Joo},
  booktitle={Proceedings of the IEEE/CVF Conference on Computer Vision and Pattern Recognition (CVPR)},
  year={2023}
}

@inproceedings{dvis,
  title={{DVIS}: Decoupled Video Instance Segmentation Framework},
  author={Zhang, Tao and Tian, Xingye and Wu, Yu and Ji, Shunping and Wang, Xuebo and Zhang, Yuan and Wan, Pengfei},
  booktitle={Proceedings of the IEEE/CVF International Conference on Computer Vision (ICCV)},
  year={2023}
}

@article{dvispp,
  title={{DVIS++}: Improved Decoupled Framework for Universal Video Segmentation},
  author={Zhang, Tao and Tian, Xingye and Zhou, Yikang and Ji, Shunping and Wang, Xuebo and Tao, Xin and Zhang, Yuan and Wan, Pengfei and Wang, Zhongyuan and Wu, Yu},
  journal={arXiv preprint arXiv:2312.13305},
  year={2023}
}

@inproceedings{syncvis,
  title={{SyncVIS}: Synchronized Video Instance Segmentation},
  author={Zheng, Rongkun and Qi, Lu and Chen, Xi and Wang, Yi and Wang, Kun and Qiao, Yu and Zhao, Hengshuang},
  booktitle={Advances in Neural Information Processing Systems (NeurIPS)},
  year={2024}
}

@article{lomm,
  title={{LOMM}: Latest Object Memory Management for Temporally Consistent Video Instance Segmentation},
  author={Lee, Seunghun and Seo, Jiwan and Choi, Minwoo and Han, Kiljoon and Jeong, Jaehoon and Durante, Zane and Adeli, Ehsan and Park, Sang Hyun and Im, Sunghoon},
  journal={arXiv preprint arXiv:2507.19754},
  year={2025},
  url={https://arxiv.org/abs/2507.19754}
}

@inproceedings{cavis,
  title={Context-Aware Video Instance Segmentation},
  author={Lee, Seunghun and Seo, Jiwan and Han, Kiljoon and Choi, Minwoo and Im, Sunghoon},
  booktitle={Proceedings of the IEEE/CVF International Conference on Computer Vision (ICCV)},
  pages={4507--4517},
  year={2025}
}

@article{xmem,
  title={{XMem}: Long-Term Video Object Segmentation with an Atkinson-Shiffrin Memory Model},
  author={Cheng, Ho Kei and Schwing, Alexander G.},
  journal={arXiv preprint arXiv:2207.07115},
  year={2022}
}

@inproceedings{cutie,
  title={Putting the Object Back into Video Object Segmentation},
  author={Cheng, Ho Kei and Oh, Seoung Wug and Price, Brian and Lee, Joon-Young and Schwing, Alexander},
  booktitle={Proceedings of the IEEE/CVF Conference on Computer Vision and Pattern Recognition (CVPR)},
  year={2024}
}

@inproceedings{sam2,
  title={{SAM} 2: Segment Anything in Images and Videos},
  author={Ravi, Nikhila and Gabeur, Valentin and Hu, Yuan-Ting and Hu, Ronghang and Ryali, Chaitanya and Ma, Tengyu and Khedr, Haitham and R{\"a}dle, Roman and Rolland, Chlo{\'e} and Gustafson, Laura and Mintun, Eric and Pan, Junting and Alwala, Kalyan Vasudev and Carion, Nicolas and Wu, Chao-Yuan and Girshick, Ross B. and Doll{\'a}r, Piotr and Feichtenhofer, Christoph},
  booktitle={International Conference on Learning Representations (ICLR)},
  year={2025},
  url={https://openreview.net/forum?id=Ha6RTeWMd0}
}

@inproceedings{sam2long,
  title={{SAM2Long}: Enhancing {SAM} 2 for Long Video Segmentation with a Training-Free Memory Tree},
  author={Ding, Shuangrui and Qian, Rui and Dong, Xiaoyi and Zhang, Pan and Zang, Yuhang and Cao, Yuhang and Guo, Yuwei and Lin, Dahua and Wang, Jiaqi},
  booktitle={Proceedings of the IEEE/CVF International Conference on Computer Vision (ICCV)},
  pages={13614--13624},
  year={2025},
  doi={10.1109/ICCV51701.2025.01264},
  url={https://doi.org/10.1109/ICCV51701.2025.01264}
}

@article{openworld_video_segmentation,
  title={Open-World Video Segmentation},
  author={Su, Qing and Li, Kaiyang and Zhuang, Yuan and Miao, Fei and Ji, Shihao},
  journal={arXiv preprint arXiv:2606.15632},
  year={2026}
}

@inproceedings{videnovic2025dam4sam,
  author={Videnovic, Jovana and Lukezic, Alan and Kristan, Matej},
  title={A Distractor-Aware Memory for Visual Object Tracking with {SAM2}},
  booktitle={Proceedings of the IEEE/CVF Conference on Computer Vision and Pattern Recognition (CVPR)},
  pages={24255--24264},
  year={2025},
  doi={10.1109/CVPR52734.2025.02259},
  url={https://openaccess.thecvf.com/content/CVPR2025/html/Videnovic_A_Distractor-Aware_Memory_for_Visual_Object_Tracking_with_SAM2_CVPR_2025_paper.html}
}

@inproceedings{dai2020ltmu,
  author={Dai, Kenan and Zhang, Yunhua and Wang, Dong and Li, Jianhua and Lu, Huchuan and Yang, Xiaoyun},
  title={High-Performance Long-Term Tracking With Meta-Updater},
  booktitle={Proceedings of the IEEE/CVF Conference on Computer Vision and Pattern Recognition (CVPR)},
  pages={6297--6306},
  year={2020},
  doi={10.1109/CVPR42600.2020.00633},
  url={https://openaccess.thecvf.com/content_CVPR_2020/html/Dai_High-Performance_Long-Term_Tracking_With_Meta-Updater_CVPR_2020_paper.html}
}

@inproceedings{ctvis,
  author={Ying, Kaining and Zhong, Qing and Mao, Weian and Wang, Zhenhua and Chen, Hao and Wu, Lin Yuanbo and Liu, Yifan and Fan, Chengxiang and Zhuge, Yunzhi and Shen, Chunhua},
  title={{CTVIS}: Consistent Training for Online Video Instance Segmentation},
  booktitle={Proceedings of the IEEE/CVF International Conference on Computer Vision (ICCV)},
  pages={899--908},
  year={2023}
}

@inproceedings{dvisdaq,
  author={Zhou, Yikang and Zhang, Tao and Ji, Shunping and Yan, Shuicheng and Li, Xiangtai},
  title={{DVIS-DAQ}: Improving Video Segmentation via Dynamic Anchor Queries},
  booktitle={Proceedings of the European Conference on Computer Vision (ECCV)},
  year={2024}
}

@inproceedings{stm,
  author={Oh, Seoung Wug and Lee, Joon-Young and Xu, Ning and Kim, Seon Joo},
  title={Video Object Segmentation Using Space-Time Memory Networks},
  booktitle={Proceedings of the IEEE/CVF International Conference on Computer Vision (ICCV)},
  pages={9226--9235},
  year={2019}
}

@inproceedings{stcn,
  author={Cheng, Ho Kei and Tai, Yu-Wing and Tang, Chi-Keung},
  title={Rethinking Space-Time Networks with Improved Memory Coverage for Efficient Video Object Segmentation},
  booktitle={Advances in Neural Information Processing Systems (NeurIPS)},
  volume={34},
  year={2021}
}

@inproceedings{aot,
  author={Yang, Zongxin and Wei, Yunchao and Yang, Yi},
  title={Associating Objects with Transformers for Video Object Segmentation},
  booktitle={Advances in Neural Information Processing Systems (NeurIPS)},
  volume={34},
  year={2021}
}

@inproceedings{deaot,
  author={Yang, Zongxin and Yang, Yi},
  title={Decoupling Features in Hierarchical Propagation for Video Object Segmentation},
  booktitle={Advances in Neural Information Processing Systems (NeurIPS)},
  volume={35},
  year={2022}
}

@article{mindthegap,
  title={Mind the Gap: Disentangling Performance Bottlenecks in Video Instance Segmentation},
  author={Hamdi, Danial and Ayar, Fardin and Javanmardi, Mahdi},
  journal={arXiv preprint arXiv:2606.07394},
  year={2026},
  url={https://arxiv.org/abs/2606.07394}
}

@article{sam3,
  title={{SAM} 3: Segment Anything with Concepts},
  author={Carion, Nicolas and Gustafson, Laura and Hu, Yuan-Ting and Debnath, Shoubhik and Hu, Ronghang and Suris, Didac and Ryali, Chaitanya and Alwala, Kalyan Vasudev and Khedr, Haitham and Huang, Andrew and Lei, Jie and Ma, Tengyu and Guo, Baishan and Kalla, Arpit and Marks, Markus and Greer, Joseph and Wang, Meng and Sun, Peize and R{\"a}dle, Roman and Afouras, Triantafyllos and Mavroudi, Effrosyni and Xu, Katherine and Wu, Tsung-Han and Zhou, Yu and Momeni, Liliane and Hazra, Rishi and Ding, Shuangrui and Vaze, Sagar and Porcher, Francois and Li, Feng and Li, Siyuan and Kamath, Aishwarya and Cheng, Ho Kei and Doll{\'a}r, Piotr and Ravi, Nikhila and Saenko, Kate and Zhang, Pengchuan and Feichtenhofer, Christoph},
  journal={arXiv preprint arXiv:2511.16719},
  year={2025},
  url={https://arxiv.org/abs/2511.16719}
}

@article{sam3dms,
  title={{SAM3-DMS}: Decoupled Memory Selection for Multi-target Video Segmentation of {SAM3}},
  author={Shen, Ruiqi and Liu, Chang and Ding, Henghui},
  journal={arXiv preprint arXiv:2601.09699},
  year={2026},
  url={https://arxiv.org/abs/2601.09699}
}

@article{mosam,
  title={{MoSAM}: Motion-Guided Segment Anything Model with Spatial-Temporal Memory Selection},
  author={Yang, Qiushi and Yao, Yuan and Cui, Miaomiao and Bo, Liefeng},
  journal={arXiv preprint arXiv:2505.00739},
  year={2025},
  url={https://arxiv.org/abs/2505.00739}
}

@article{dual_state_slot_attention,
  title={Dual-State Slot Attention: Decoupling Appearance and Identity for Video Object-Centric Learning},
  author={Tran, Sieu and Nguyen, Duc and Vo, Hao and Vo, Khoa and Le, Ngan},
  journal={arXiv preprint arXiv:2606.12601},
  year={2026},
  url={https://arxiv.org/abs/2606.12601}
}

@article{temporal_slot_activation,
  title={{TSA}: Temporal Slot Activation for Persistent Object-Centric Video Representation},
  author={Nguyen, Duc and Tran, Sieu and Vo, Hao and Vo, Khoa and Nguyen, Duy Minh Ho and Bui, Nghi D. Q. and Nguyen, Anh and Mai, Long and Le, Ngan},
  journal={arXiv preprint arXiv:2606.13714},
  year={2026},
  url={https://arxiv.org/abs/2606.13714}
}

@inproceedings{wang2026stc,
  title={Accelerating Streaming Video Large Language Models via Hierarchical Token Compression},
  author={Wang, Yiyu and Liu, Xuyang and Gui, Xiyan and Lin, Xinying and Yang, Boxue and Liao, Chenfei and Chen, Tailai and Zhang, Linfeng},
  booktitle={Proceedings of the IEEE/CVF Conference on Computer Vision and Pattern Recognition (CVPR)},
  year={2026},
  url={https://arxiv.org/abs/2512.00891}
}

@article{wen2025ai4service,
  title={{AI} for Service: Proactive Assistance with {AI} Glasses},
  author={Wen, Zichen and Wang, Yiyu and Liao, Chenfei and Yang, Boxue and Li, Junxian and Liu, Weifeng and He, Haocong and Feng, Bolong and Liu, Xuyang and Lyu, Yuanhuiyi and Zheng, Xu and Hu, Xuming and Zhang, Linfeng},
  journal={arXiv preprint arXiv:2510.14359},
  year={2025},
  url={https://arxiv.org/abs/2510.14359}
}

@article{li2026videococo,
  title={{VideoCoCo}: Code-as-{CoT} for Physically-Consistent Video Generation via an Agentic Dual-Engine System},
  author={Li, Haodong and Ren, Tianfei and Ma, Xiaoxiao and Qing, Chunmei and Fang, Zhen and He, Sipeng and Guo, Ziyu and Wu, Haoyu and Tian, Juanxi and Zou, Yihang and An, Ruichuan and Jiang, Dongzhi and Yang, Boxue and Xie, Ji and Huang, Xu and Yan, Wenhao and Zou, Jialv and Yue, Zhengrong and Luo, Yaxin and Li, Xiaotong and Wang, Yuzhu and Ye, Junyan and Zhao, Jinjing and Chen, Zehui and Chen, Lin and Yan, Renye and Zhao, Feng and Heng, Pheng-Ann},
  journal={arXiv preprint arXiv:2607.27380},
  year={2026},
  url={https://arxiv.org/abs/2607.27380}
}

@article{wen2026evostreaming,
  title={{EvoStreaming}: Your Offline Video Model Is a Natively Streaming Assistant},
  author={Wen, Zichen and Yang, Boxue and Ke, Junlong and Huang, Jiajie and Liao, Chenfei and Wang, Junxi and Liu, Xuyang and Zhang, Linfeng},
  journal={arXiv preprint arXiv:2605.10343},
  year={2026},
  url={https://arxiv.org/abs/2605.10343}
}

@article{li2026dvlt,
  title={{4DVLT}: Dynamic Scene Understanding with Worldline-Centered Vision-Language Tracking},
  author={Li, Chaoyue and Yang, Boxue and Zhou, Shengyao and Wu, Haoyang and Qian, Rui and Zhang, Linfeng},
  journal={arXiv preprint arXiv:2606.22631},
  year={2026},
  url={https://arxiv.org/abs/2606.22631}
}

@inproceedings{savi,
  title={Conditional Object-Centric Learning from Video},
  author={Kipf, Thomas and Elsayed, Gamaleldin Fathy and Mahendran, Aravindh and Stone, Austin and Sabour, Sara and Heigold, Georg and Jonschkowski, Rico and Dosovitskiy, Alexey and Greff, Klaus},
  booktitle={International Conference on Learning Representations (ICLR)},
  year={2022},
  url={https://openreview.net/forum?id=aD7uesX1GF_}
}

@inproceedings{vaswani2017attention,
  title={Attention Is All You Need},
  author={Vaswani, Ashish and Shazeer, Noam and Parmar, Niki and
          Uszkoreit, Jakob and Jones, Llion and Gomez, Aidan N. and
          Kaiser, Lukasz and Polosukhin, Illia},
  booktitle={Advances in Neural Information Processing Systems},
  volume={30},
  year={2017}
}

@inproceedings{cho2014learning,
  title={Learning Phrase Representations using {RNN} Encoder--Decoder for
         Statistical Machine Translation},
  author={Cho, Kyunghyun and van Merri{\"e}nboer, Bart and Gulcehre, Caglar
          and Bahdanau, Dzmitry and Bougares, Fethi and Schwenk, Holger
          and Bengio, Yoshua},
  booktitle={Proceedings of the 2014 Conference on Empirical Methods in
             Natural Language Processing},
  pages={1724--1734},
  year={2014},
  publisher={Association for Computational Linguistics}
}

@inproceedings{hong2023lvos,
  author={Hong, Lingyi and Chen, Wenchao and Liu, Zhongying and Zhang, Wei and Guo, Pinxue and Chen, Zhaoyu and Zhang, Wenqiang},
  title={{LVOS}: A Benchmark for Long-term Video Object Segmentation},
  booktitle={Proceedings of the IEEE/CVF International Conference on Computer Vision (ICCV)},
  pages={13434--13446},
  year={2023},
  doi={10.1109/ICCV51070.2023.01240}
}

@article{hong2024lvosv2,
  author={Hong, Lingyi and Liu, Zhongying and Chen, Wenchao and Tan, Chenzhi and Feng, Yuang and Zhou, Xinyu and Guo, Pinxue and Li, Jinglun and Chen, Zhaoyu and Gao, Shuyong and Zhang, Wei and Zhang, Wenqiang},
  title={{LVOS}: A Benchmark for Large-Scale Long-Term Video Object Segmentation},
  journal={IEEE Transactions on Pattern Analysis and Machine Intelligence},
  volume={48},
  number={1},
  pages={946--961},
  year={2026},
  doi={10.1109/TPAMI.2025.3611020}
}

@book{boyd2004convex,
  author={Boyd, Stephen and Vandenberghe, Lieven},
  title={Convex Optimization},
  publisher={Cambridge University Press},
  year={2004},
  url={https://web.stanford.edu/~boyd/cvxbook/}
}

@inproceedings{perazzi2016davis,
  author={Perazzi, Federico and Pont-Tuset, Jordi and McWilliams, Brian and Van Gool, Luc and Gross, Markus H. and Sorkine-Hornung, Alexander},
  title={A Benchmark Dataset and Evaluation Methodology for Video Object Segmentation},
  booktitle={Proceedings of the IEEE Conference on Computer Vision and Pattern Recognition (CVPR)},
  pages={724--732},
  year={2016},
  doi={10.1109/CVPR.2016.85}
}
\bibliographystyle{iclr2027_conference}

\clearpage
\appendix
\captionsetup{font=footnotesize,skip=4pt,justification=raggedright,singlelinecheck=false}
\setlength{\intextsep}{8pt plus 1pt minus 1pt}
\section{Mathematical Properties of State Reasoning}
\label{sec:mathematical_properties}

This section analyzes the association, verification, and state-transition rules
in Section~\ref{sec:method}, together with their refinement supervision. The
results characterize how the rules handle competing and uncertain evidence;
segmentation performance is evaluated empirically in Section~\ref{sec:experiments}.

\begingroup
\setlength{\parskip}{2pt plus 1pt}
\renewcommand{\theequation}{\thesection.\arabic{equation}}
\renewcommand{\theHequation}{\thesection.\arabic{equation}}
\setcounter{equation}{0}

\subsection{Sparse Association and Competition}
\label{sec:math_association}

\paragraph{Notation and assumptions.}
For a fixed frame, let $A_i=\mathcal{N}_i\cup\{\varnothing\}$ be the finite
legal candidate set of state $i$. The null candidate is always present. We
assume finite legal logits, $\tau>0$, and
$\lambda=\lambda_{\mathrm{cmp}}\ge0$. Illegal edges have probability zero.
For the conditional analysis below, hold $A_i$ and the compatibility logits
$e_{ij}$ fixed, and write
\begin{equation}
 p_{ij}(c)=\frac{\exp(e_{ij}-\lambda c_j)}
 {\sum_{k\in A_i}\exp(e_{ik}-\lambda c_k)},\qquad j\in A_i,
 \quad c_\varnothing=0.
 \label{eq:app_distribution}
\end{equation}
The positive denominator ensures $p_{ij}\ge0$ and $\sum_jp_{ij}=1$.
If $\mathcal{N}_i$ is empty, $p_{i\varnothing}=1$: the rule remains defined
without forcing a real association. Consequently, the matched evidence
$\bar u_i=\sum_jp_{ij}u_j$ lies in the convex hull of the legal observation
features, including the learned null feature.

\paragraph{Proposition 1 (effect of competition).}
For any $j,k\in A_i$, the association odds satisfy
\begin{equation}
 \frac{p_{ij}(c)}{p_{ik}(c)}
 =\frac{p_{ij}(0)}{p_{ik}(0)}e^{-\lambda(c_j-c_k)}.
 \label{eq:app_odds}
\end{equation}
In particular, a real candidate's odds relative to null are multiplied by
$e^{-\lambda c_j}$. Let $M_j(c)=\sum_i p_{ij}(c)$. Increasing only the
penalty of real candidate $j$ gives
\begin{equation}
 \frac{\partial M_j}{\partial c_j}
 =-\lambda\sum_{i:j\in A_i}p_{ij}(1-p_{ij})\le0,
 \qquad
 \frac{\partial p_{i\varnothing}}{\partial c_j}
 =\lambda p_{i\varnothing}p_{ij}\ge0.
 \label{eq:app_competition_derivative}
\end{equation}
Thus, componentwise increases in real-candidate penalties cannot decrease
the null probability when the compatibility logits are fixed.

\emph{Proof.}
Taking the ratio of two terms in Equation~\ref{eq:app_distribution} cancels
their common normalizer and yields Equation~\ref{eq:app_odds}. Differentiation
with respect to a legal real-candidate penalty gives
$\partial p_{ik}/\partial c_j
=-\lambda p_{ik}(\mathbf{1}\{k=j\}-p_{ij})$.
Taking $k=j$ and summing over states proves the first derivative; taking
$k=\varnothing$ proves the second. An illegal edge contributes zero.
Integrating the nonnegative null derivatives along a componentwise increasing
penalty path proves the final statement.\hfill$\square$

This result isolates the contribution of the competition term. Verify also
changes the contextualized logits, so it does not imply that every column
mass decreases between the two reasoning steps. Soft competition shapes
relative preferences; the assignment-based output adapter separately enforces
at most one retained claim per real observation by selecting a single
highest-ranked claimant (with ties resolved to one claimant). Null is exempt
from both the excess-demand penalty and this exclusivity rule.

\paragraph{Decision features.}

The entropy and occupancy terms in Equation~\ref{eq:decision_feature}
describe different aspects of an association. With natural logarithms and
$0\log0=0$, $0\le\mathcal{E}(p_i)\le\log|A_i|$. The lower bound follows
from $-p\log p\ge0$; the upper bound uses KL nonnegativity
\citep[Example~3.19]{boyd2004convex}:
$\mathrm{KL}(p_i\|\mathrm{Unif}(A_i))=\log|A_i|-\mathcal{E}(p_i)\ge0$.
Padding illegal edges with zeros puts all rows in the same candidate space,
where the occupancy feature has the exact decomposition
\begin{equation}
 \rho_i=\|p_i\|_2^2+\sum_{k\ne i}\langle p_i,p_k\rangle,
 \qquad
 \sum_i\rho_i=\sum_{j\ne\varnothing}M_j^2+M_\varnothing^2.
 \label{eq:app_occupancy}
\end{equation}
Indeed, substituting $M_j=\sum_kp_{kj}$ into $\rho_i=\sum_jp_{ij}M_j$
and separating $k=i$ gives the first equality; exchanging the two sums gives
the second. Occupancy therefore combines within-state concentration with
cross-state overlap, whereas entropy depends only on the individual row.
The null contribution provides context about unmatched states; it is not a
real-observation conflict, consistent with $c_\varnothing=0$.

\subsection{Controlled Refinement of Transition Decisions}

\paragraph{Proposition 2 (action-distribution perturbation).}
Let $a$ be the proposed action logits and
$d=\alpha_a\odot(\widetilde a^2-a)$ their gated correction, both finite. Set
$\pi^1=\operatorname{softmax}(a)$,
$\pi^2=\operatorname{softmax}(a+d)$, and $D=\|d\|_\infty$. Then
\begin{equation}
 \|\pi^2-\pi^1\|_1\le\min\{2,D\},\qquad
 \mathrm{KL}(\pi^1\|\pi^2)\le\tfrac12D^2.
 \label{eq:app_refinement_bound}
\end{equation}
No sign or interval constraint on $\alpha_a$ is needed. In particular,
$D\le\|\alpha_a\|_\infty\|\widetilde a^2-a\|_\infty$.

\emph{Proof.}
Consider $\pi(t)=\operatorname{softmax}(a+td)$ for $t\in[0,1]$.
Writing $\mu_t=\sum_k\pi_k(t)d_k$, differentiation gives
$\pi'_k(t)=\pi_k(t)(d_k-\mu_t)$. Since $|d_k|\le D$,
\begin{equation}
 \|\pi'(t)\|_1
 =\mathbb{E}_{\pi(t)}|d-\mu_t|
 \le\sqrt{\operatorname{Var}_{\pi(t)}(d)}\le D.
\end{equation}
Integration yields the $D$ bound, while two probability distributions are
always at most 2 apart in $\ell_1$. For the log-sum-exp function
\citep[Section~3.1.5]{boyd2004convex},
$f(t)=\log\sum_k\exp(a_k+td_k)$, we have
$f'(t)=\mu_t$ and $f''(t)=\operatorname{Var}_{\pi(t)}(d)\le D^2$. Hence
\begin{equation}
 \mathrm{KL}(\pi^1\|\pi^2)
 =f(1)-f(0)-f'(0)
 =\int_0^1(1-t)f''(t)\,dt\le\tfrac12D^2.
\end{equation}
This proves both claims.\hfill$\square$

The bound explains how small gated logit corrections preserve a proposal's
action distribution. It concerns the size of a correction, not its accuracy.
For association probabilities, the change in the competition penalty must
also be included in the effective logit correction.

\subsection{Selective Admission to Persistent Memory}

\paragraph{Proposition 3 (quadratic attenuation of uncertain evidence).}
For a single state, let $a=\pi^2_{\mathrm{update}}+\pi^2_{\mathrm{revive}}$
and $s=1-p^2_\varnothing$ denote the write-action probability and total real
association mass. Define the conditional concentration
\begin{equation}
 \kappa=\begin{cases}
 \displaystyle\max_{j\in\mathcal{N}}p_j^2/s,&s>0,\\
 0,&s=0.
 \end{cases}
 \qquad
 g=a\kappa s^2,\quad 0\le g\le as^2\le s^2\le1.
 \label{eq:app_gate_factorization}
\end{equation}
Here the maximum over an empty real-candidate set is defined as zero.
For any norm, the state transition in Equation~\ref{eq:state_transition}
satisfies
\begin{equation}
 \|z_t-z_{t-1}\|
 =g\|\hat z_t-z_{t-1}\|
 \le as^2\|\hat z_t-z_{t-1}\|.
 \label{eq:app_step_bound}
\end{equation}
In particular, $p^2_\varnothing\ge1-\varepsilon$, with
$\varepsilon\in[0,1]$, implies $g\le\varepsilon^2$.

\emph{Proof.}
Both action and association probabilities are normalized, so $a,s\in[0,1]$.
For $s>0$, the maximum real probability is $\kappa s$ with
$\kappa\in[0,1]$; substituting this into the write rule gives
$g=a\kappa s^2$. If $s=0$, the real probabilities and the gate are zero,
so the same identity holds. Subtracting $z_{t-1}$ from the update and using
norm homogeneity proves Equation~\ref{eq:app_step_bound}.
Finally, $s\le\varepsilon$ gives the claimed attenuation.\hfill$\square$

The factors isolate action preference, conditional concentration, and
real-evidence mass. With $K\ge1$ real candidates and $s>0$,
$1/K\le\kappa\le1$ because the conditional probabilities sum to one.
The update is convex: $g=0$ preserves the identity feature, though lifecycle
metadata may advance. For example, $p^2_\varnothing\ge0.9$ gives $g\le0.01$,
limiting feature movement to one percent of the candidate displacement.
This is a relative bound; an absolute bound additionally requires bounding
$\|\hat z_t-z_{t-1}\|$.

\paragraph{Corollary 3.1 (retention along a state trajectory).}
Consider one identity over $L$ frames without slot reinitialization. Define
$R_{u:v}=\prod_{r=u}^{v}(1-g_r)$, with an empty product equal to one.
For the realized gates and candidate states,
\begin{equation}
 z_L=R_{1:L}z_0+\sum_{t=1}^{L}g_tR_{t+1:L}\hat z_t,
 \qquad R_{1:L}+\sum_{t=1}^{L}g_tR_{t+1:L}=1.
 \label{eq:app_unrolling}
\end{equation}
All coefficients are nonnegative. If
$p^2_{t,\varnothing}\ge1-\varepsilon_t$ with $\varepsilon_t\in[0,1]$
at each frame, then
\begin{equation}
 1-R_{1:L}
 \le1-\prod_{t=1}^{L}(1-\varepsilon_t^2)
 \le\min\left\{1,\sum_{t=1}^{L}\varepsilon_t^2\right\}.
 \label{eq:app_retention_bound}
\end{equation}

\emph{Proof.}
Repeated substitution of the single-write recurrence gives the first
identity. The second follows by telescoping
$g_tR_{t+1:L}=R_{t+1:L}-R_{t:L}$. Proposition 3 gives
$g_t\le\varepsilon_t^2$, hence
$R_{1:L}\ge\prod_t(1-\varepsilon_t^2)$. The final inequality follows
by induction from $1-(1-x)(1-y)=x+y-xy\le x+y$ for $x,y\in[0,1]$,
together with $0\le\prod_t(1-\varepsilon_t^2)\le1$.\hfill$\square$

For a uniform bound $\varepsilon_t\le\varepsilon\le1$, the retained coefficient
is at least $(1-\varepsilon^2)^L$. Moreover, if
$\|z_0\|,\|\hat z_t\|\le B$ throughout the interval, convexity gives
$\|z_L\|\le B$. These statements hold for the realized trajectory even
when gates and candidates depend on earlier states. The coefficients are
algebraic mixture weights, not derivatives of the trajectory with respect to
$z_0$; a contraction claim would require additional control of those
dependencies. Propose and Verify contribute only through the final gate and
candidate, since neither intermediate step writes to persistent memory.

\subsection{Meaning of Refinement Supervision}

For a nonempty valid-state set $\mathcal{V}$, put
$d_i=[p^1_{i,y_i}-p^2_{i,y_i}]_+$ and
$n=|\mathcal{V}|$. Since every $d_i$ is nonnegative,
$\mathcal{L}_{\mathrm{ref}}=0$ if and only if
$p^2_{i,y_i}\ge p^1_{i,y_i}$ for every $i\in\mathcal{V}$.
More generally, for any $\epsilon>0$, let
$B_\epsilon=\{i\in\mathcal{V}:p^1_{i,y_i}-p^2_{i,y_i}>\epsilon\}$.
Then
\begin{equation}
 \frac{|B_\epsilon|}{n}
 \le\min\{1,\mathcal{L}_{\mathrm{ref}}/\epsilon\}.
 \label{eq:app_supervision_bound}
\end{equation}
To see this, sum $\epsilon\mathbf{1}\{i\in B_\epsilon\}\le d_i$ over
the valid states and divide by $n\epsilon$. Thus, the refinement loss
controls the proportion of states whose target-association probability
deteriorates by more than a prescribed amount on the supervised samples.
The transition-advantage target in Equation~\ref{eq:transition_advantage}
separately specifies whether a candidate improves the chosen continuation
quality relative to retaining the state. Neither the joint objective nor
the algebraic properties above assume exact optimization or perfect
advantage prediction; their empirical effect is assessed by the reported
comparisons and controlled analyses.

\endgroup

\section{Implementation Details}
\label{sec:implementation_details}

\subsection{Host Interface}
POSReasoner consumes frozen trajectory scores, masks, identity descriptors, and
their temporal summaries. It does not change host weights or request additional
image-level proposals. For every matched comparison, the host checkpoint, input
resolution, candidate export, and evaluator are held fixed. We preserve each
host's native output budget unless a larger candidate export is explicitly
reported as part of that host configuration.

\subsection{Training and Evaluation}
\paragraph{VIS reasoner.}
The VIS instantiation uses a 96-dimensional hidden space, four attention heads,
and two relation layers, trained for 180 epochs with AdamW, a learning rate of
$1.5\times10^{-3}$, and weight decay of $2\times10^{-4}$. Five video-grouped
folds select the executor from out-of-fold predictions, and their models are
averaged at inference. Candidate budgets are fixed before validation. For hosts
without frame-wise identity sequences, a compact readout uses their state and
relation descriptors under the same protocol.

\paragraph{VOS reasoner.}
The VOS instantiation replays frozen identity-indexed masks and constructs events
from visibility, geometry, and cross-trajectory changes. Persistent-state,
reactivation, and sparse multi-trajectory policies are fitted or calibrated on
video-disjoint training splits, then frozen before validation. Readout videos
remain disjoint from calibration and holdout evaluation.

\paragraph{Selection and evaluation.}
No validation annotation is read during inference; only the official evaluator
and the post-hoc analysis in Table~\ref{tab:reappearance_appendix} load it. Main
benchmarks are single runs of a fixed host--reasoner pair, while the OVIS
mechanism study repeats seeds 20260728, 20260729, and 20260730. Hyperparameters
and executor choices use grouped out-of-fold training performance and are fixed
before validation.

\subsection{Controlled Comparisons and Efficiency}

The controls use the same frozen candidates and training protocol.

\begingroup
\setlength{\columnsep}{12pt}
\setlength{\intextsep}{3pt}
\begin{wraptable}{r}{0.56\textwidth}
    \centering
    \captionsetup{font=footnotesize,width=\linewidth,skip=4pt}
    \caption{Causal controls on OVIS with the same frozen
    candidates and training protocol.}
    \label{tab:vis_reasoning_appendix}
    \begingroup
    \footnotesize
    \renewcommand{\arraystretch}{1.06}
    \setlength{\tabcolsep}{2.3pt}
    \begin{tabularx}{\linewidth}{@{}Xrrrr@{}}
            \toprule
            \rowcolor{black!5}
            Variant & AP & $\Delta$AP & AP50 & AP75 \\
            \midrule
            Frozen host & 34.63 & -- & 59.94 & 33.80 \\
            Host cues only & 34.82 & +0.18 & 60.21 & 34.09 \\
            Shuffled state & 34.67 & +0.04 & 59.95 & 33.94 \\
            State, no graph & 36.55 & +1.92 & 62.88 & 35.61 \\
            State, one step & 36.13 & +1.50 & 62.01 & 35.27 \\
            State, shared two-step & 36.97 & +2.33 & 63.00 & 36.17 \\
            \rowcolor{POSRbg}
            \textcolor{POSRaccent}{\textbf{Full POSReasoner}}
            & \textbf{37.31} & \gain{+2.68}
            & \textbf{63.20} & \textbf{36.98} \\
            \bottomrule
    \end{tabularx}
    \endgroup

\end{wraptable}
\noindent
Host cues alone and shuffled states remain near the frozen host
(Table~\ref{tab:vis_reasoning_appendix}). Intact states reach 36.55 AP even
without a graph; the full model reaches 37.31 AP, supporting structured
interaction beyond state retention. Shared two-step reasoning also exceeds
one-step reasoning (36.97 versus 36.13 AP), consistent with the benefit of
revisiting initial decisions. These controls support using identity-specific
history together with structured interaction. Across three seeds, the full model achieves
$37.203\pm0.099$ AP, a $2.570\pm0.099$ gain.
\par

\begin{wraptable}{l}{0.50\textwidth}
    \centering
    \captionsetup{font=footnotesize,width=\linewidth,skip=4pt}
    \caption{Five-member VIS reasoner efficiency on one A100,
    batch size 1.}
    \label{tab:efficiency}
    \begingroup
    \footnotesize
    \renewcommand{\arraystretch}{1.08}
    \setlength{\tabcolsep}{3pt}
    \begin{tabularx}{\linewidth}{@{}Xrrr@{}}
            \toprule
            \rowcolor{black!5}
            Candidates & Params & Latency & Memory \\
            \midrule
            10 & 0.90M & 5.80 ms & 12.71 MB \\
            20 & 0.90M & 5.66 ms & 12.99 MB \\
            40 & 0.90M & 5.64 ms & 14.03 MB \\
            \bottomrule
    \end{tabularx}
    \endgroup

\end{wraptable}
\noindent
The five-member VIS reasoner uses 0.90M parameters across budgets of 10--40
candidates (Table~\ref{tab:efficiency}). Median latency remains within
5.64--5.80 ms, while peak allocated CUDA memory increases from 12.71 to
14.03 MB. These measurements cover the reasoner, with latency measured over
500 runs after 100 warm-ups.
\par
Experiments use NVIDIA A100-SXM4 80\,GB GPUs and Intel Xeon Platinum 8369B CPUs
under Ubuntu 24.04, with Python 3.10.20, PyTorch 2.4.1, CUDA 12.1, and NumPy
1.26.4. Independent folds and dataset--host pairs run in parallel; each frozen
replay needs one GPU.
\par
\endgroup

\begin{minipage}{\textwidth}
\section{Reappearance Analysis}
\label{sec:reappearance_analysis}
\noindent
\begin{minipage}[t]{0.51\textwidth}
\vspace{0pt}
Figure~\ref{fig:additional_lvos_qualitative} illustrates identity recovery, while
Table~\ref{tab:reappearance_appendix} groups reappearance events by absence
duration. These post-hoc diagnostics distinguish long-term recovery from
ordinary frame-to-frame matching and do not affect policy selection.
Intermediate gaps show the clearest gains when local identity is ambiguous
but persistent evidence remains.

Gaps are measured in absent frames. The table lists region similarity, event
and video counts, and paired bootstrap intervals over videos. H/T/D counts
events that improve, remain unchanged, or degrade.
\par
\end{minipage}\hfill
\begin{minipage}[t]{0.46\textwidth}
    \vspace{0pt}
    \centering
    \includegraphics[width=\linewidth]{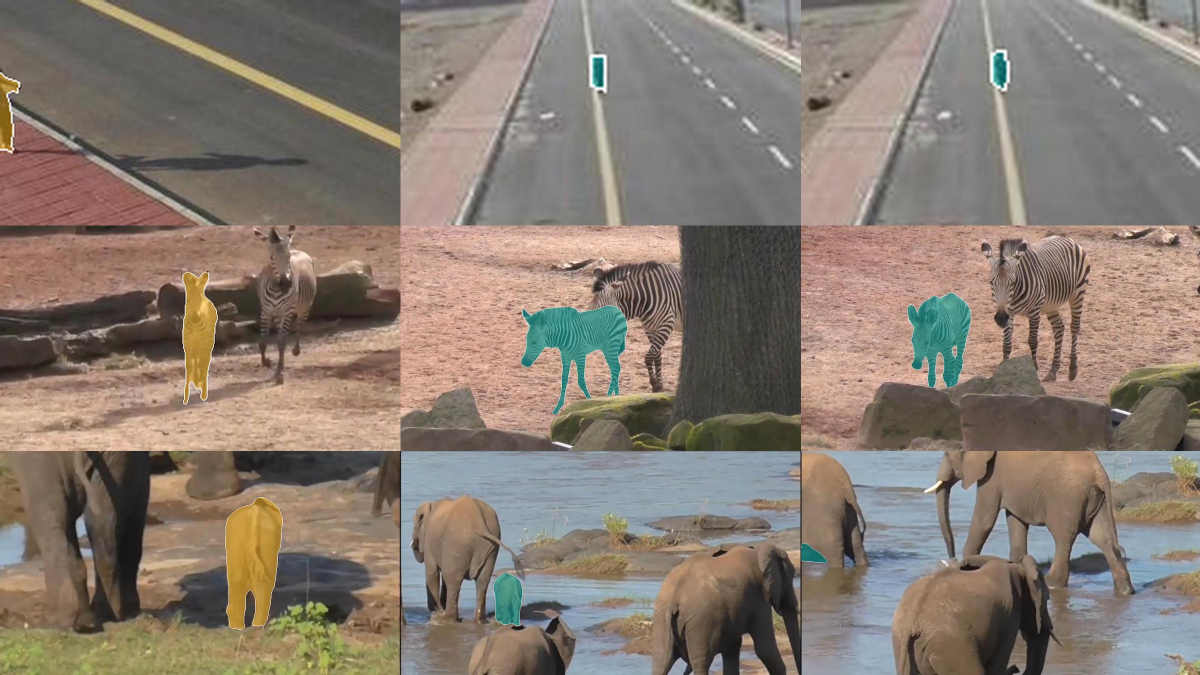}
    \captionsetup{font=footnotesize,skip=4pt,width=\linewidth,hypcap=false}
    \captionof{figure}{LVOS v1 reappearance, similar-object interaction,
    and crowded motion. Yellow: references; teal: unmodified POSReasoner outputs.}
    \label{fig:additional_lvos_qualitative}
\end{minipage}

\par\smallskip
\begin{table}[H]
    \centering
    \captionsetup{font=footnotesize,skip=4pt,width=0.96\textwidth}
    \caption{Post-reappearance results by absence duration, with paired
    differences and 95\% bootstrap confidence intervals.}
    \label{tab:reappearance_appendix}
    \begingroup
    \footnotesize
    \setlength{\tabcolsep}{3.4pt}
    \renewcommand{\arraystretch}{1.08}
    \begin{tabularx}{0.96\textwidth}{@{}l*{3}{>{\raggedleft\arraybackslash}X}rrrr@{}}
        \toprule
        \rowcolor{black!5}
        Dataset & Gap & Events & Videos & Host $\mathcal{J}$ &
        POSR $\mathcal{J}$ & $\Delta\mathcal{J}$ [95\% CI] & H/T/D \\
        \midrule
        \multirow{4}{*}{LVOS v1}
        & 5--10   & 45  & 16 & 70.6 & 74.9 & $+4.3$ [$-2.3$, $16.3$] & 22/9/14 \\
        & 11--30  & 34  & 15 & 51.1 & 64.2
        & \cellcolor{POSRbg}\textcolor{POSRaccent}{\textbf{$+13.1$}}
        [$-3.1$, $29.1$] & 20/6/8 \\
        & 31--100 & 14  & 8  & 72.2 & 78.0 & $+5.9$ [$-2.7$, $19.1$] & 7/3/4 \\
        & $>100$  & 6   & 4  & 23.6 & 30.6 & $+7.1$ [$-10.1$, $46.9$] & 1/2/3 \\
        \midrule
        \multirow{4}{*}{LVOS v2}
        & 5--10   & 114 & 42 & 66.9 & 65.8 & $-1.1$ [$-7.7$, $5.3$] & 53/17/44 \\
        & 11--30  & 83  & 39 & 56.6 & 61.1 & $+4.5$ [$-2.3$, $11.6$] & 38/15/30 \\
        & 31--100 & 33  & 22 & 67.1 & 69.6 & $+2.5$ [$-4.5$, $9.7$] & 17/5/11 \\
        & $>100$  & 9   & 7  & 26.2 & 24.1 & $-2.0$ [$-5.5$, $0.1$] & 1/7/1 \\
        \bottomrule
    \end{tabularx}
    \endgroup
\end{table}

\end{minipage}
\clearpage
\section{Additional Benchmark Results}
\label{sec:additional_benchmarks}
Tables~\ref{tab:ovis_full_preview} and~\ref{tab:ytvis_full_preview} extend the
OVIS and YouTube-VIS comparisons; paired runs share the host, candidate
budget, and evaluator.
\begin{table}[H]
    \centering
    \captionsetup{font=footnotesize,skip=4pt,width=0.96\textwidth}
    \caption{Complete OVIS evaluation across three backbones. Cited rows are
    published results; each uncited host is paired with POSReasoner under
    the same frozen host and candidate set.}
    \label{tab:ovis_full_preview}
    \begingroup
    \footnotesize
    \renewcommand{\arraystretch}{1.04}
    \setlength{\tabcolsep}{6pt}
    \begin{tabularx}{0.96\textwidth}{@{}X c *{3}{>{\raggedleft\arraybackslash}p{0.095\textwidth}}@{}}
        \toprule
        \rowcolor{black!5}
        Method & Backbone & AP & AP50 & AP75 \\
        \midrule
        GenVIS~\citep{genvis} & ResNet-50 & 35.8 & 60.8 & 36.2 \\
        CTVIS~\citep{ctvis} & ResNet-50 & 35.5 & 60.8 & 34.9 \\
        DVIS++~\citep{dvispp} & ResNet-50 & 37.2 & 62.8 & 37.3 \\
        DAQ~\citep{dvisdaq} & ResNet-50 & 38.7 & 65.5 & 37.6 \\
        \addlinespace[2pt]
        CTVIS & ResNet-50 & 34.6 & 59.9 & 33.8 \\
        \rowcolor{POSRbg}
        \textcolor{POSRaccent}{\textbf{CTVIS + POSReasoner}} & ResNet-50 & \textbf{37.3} & \textbf{63.2} & \textbf{37.0} \\
        DVIS++ & ResNet-50 & 36.8 & 61.3 & 37.5 \\
        \rowcolor{POSRbg}
        \textcolor{POSRaccent}{\textbf{DVIS++ + POSReasoner}} & ResNet-50 & \textbf{38.9} & \textbf{64.3} & \textbf{39.4} \\
        DAQ & ResNet-50 & 38.3 & 65.2 & 37.3 \\
        \rowcolor{POSRbg}
        \textcolor{POSRaccent}{\textbf{DAQ + POSReasoner}} & ResNet-50 & \textbf{39.9} & \textbf{66.7} & \textbf{39.1} \\
        \midrule
        MinVIS~\citep{minvis} & Swin-L & 39.4 & 61.5 & 41.3 \\
        IDOL~\citep{idol} & Swin-L & 40.0 & 63.1 & 40.5 \\
        GenVIS~\citep{genvis} & Swin-L & 45.2 & 69.1 & 48.4 \\
        DVIS~\citep{dvis} & Swin-L & 45.9 & 71.1 & 48.3 \\
        CTVIS~\citep{ctvis} & Swin-L & 46.9 & 71.5 & 47.5 \\
        LOMM~\citep{lomm} & Swin-L & 47.8 & 73.6 & 51.4 \\
        \addlinespace[2pt]
        DAQ & Swin-L & 49.6 & 76.0 & 52.8 \\
        \rowcolor{POSRbg}
        \textcolor{POSRaccent}{\textbf{DAQ + POSReasoner}} & Swin-L & \textbf{50.6} & \textbf{77.0} & \textbf{53.9} \\
        \midrule
        DVIS++~\citep{dvispp} & ViT-L & 49.6 & 72.5 & 55.0 \\
        LOMM~\citep{lomm} & ViT-L & 51.7 & 73.9 & 57.5 \\
        \addlinespace[2pt]
        DAQ & ViT-L & 53.9 & 78.9 & 58.5 \\
        \rowcolor{POSRbg}
        \textcolor{POSRaccent}{\textbf{DAQ + POSReasoner}} & ViT-L & \textbf{55.2} & \textbf{79.7} & \textbf{59.9} \\
        \bottomrule
    \end{tabularx}
    \endgroup
\end{table}

\begin{table}[H]
    \centering
    \captionsetup{font=footnotesize,skip=4pt,width=0.96\textwidth}
    \caption{Complete YouTube-VIS plug-in evaluation; 2022 baselines follow
    LOMM~\citep{lomm}. Each POSReasoner row retains its frozen host; $^*$ denotes
    offline refinement and $\dagger$ provisional results.}
    \label{tab:ytvis_full_preview}
    \begingroup
    \footnotesize
    \setlength{\tabcolsep}{2.4pt}
    \renewcommand{\arraystretch}{1.06}
    \begin{tabularx}{0.96\textwidth}{@{}X c rrr rrr rrr@{}}
        \toprule
        \rowcolor{black!5}
        \multirow{2}{*}{Method} &
        \multirow{2}{*}{Backbone} &
        \multicolumn{3}{c}{YTVIS 2019} &
        \multicolumn{3}{c}{YTVIS 2021} &
        \multicolumn{3}{c}{YTVIS 2022} \\
        \cmidrule(lr){3-5}
        \cmidrule(lr){6-8}
        \cmidrule(l){9-11}
        \rowcolor{black!5}
        & & AP & AP50 & AP75 & AP & AP50 & AP75 & AP & AP50 & AP75 \\
        \midrule
        \rowcolor{black!5}
        \multicolumn{11}{c}{\textbf{Query-based Association}} \\
        MinVIS~\citep{minvis} & R50
        & 47.4 & 69.0 & 52.1
        & 44.2 & 66.0 & 48.1
        & 23.3 & 47.9 & 19.3 \\
        VITA$^*$~\citep{vita} & R50
        & 49.8 & 72.6 & 54.5
        & 45.7 & 67.4 & 49.5
        & 32.6 & 53.9 & 39.3 \\
        GenVIS~\citep{genvis} & R50
        & 50.0 & 71.5 & 54.6
        & 47.1 & 67.5 & 51.5
        & 37.5 & 61.6 & 41.5 \\
        \rowcolor{black!5}
        \multicolumn{11}{c}{\textbf{Memory and Decoupled Tracking}} \\
        DVIS++~\citep{dvispp} & R50
        & 55.5 & 80.2 & 60.1
        & 50.0 & 72.2 & 54.5
        & 37.2 & 57.4 & 40.7 \\
        LOMM~\citep{lomm} & R50
        & 55.7 & 79.8 & 61.4
        & 50.7 & 72.9 & 56.9
        & 41.1 & 62.4 & 46.2 \\
        DVIS++~\citep{dvispp} & ViT-L
        & 67.7 & 88.8 & 75.3
        & 62.3 & 82.7 & 70.2
        & 37.5 & 53.7 & 39.4 \\
        LOMM~\citep{lomm} & ViT-L
        & 69.1 & 89.3 & 76.5
        & 65.0 & 86.0 & 72.7
        & 48.2 & 70.5 & 53.2 \\
        \midrule
        \rowcolor{black!5}
        \multicolumn{11}{c}{\textbf{Persistent Object-State Reasoning}} \\
        CTVIS & R50
        & 51.8 & 74.1 & 55.2
        & 49.5 & 72.5 & 53.6
        & 45.4 & 67.5 & 48.9 \\
        \rowcolor{POSRbg}
        \textcolor{POSRaccent}{\textbf{CTVIS + POSReasoner}} & R50
        & \textbf{52.8} & \textbf{75.2} & \textbf{56.5}
        & \textbf{50.6} & \textbf{73.7} & \textbf{55.1}
        & \textbf{47.5} & \textbf{69.7} & \textbf{51.4} \\
        \addlinespace[1.5pt]
        DVIS++ & R50
        & 55.7 & 80.8 & 59.8
        & 50.4 & 71.4 & 55.2
        & 46.3 & 66.6 & 50.5 \\
        \rowcolor{POSRbg}
        \textcolor{POSRaccent}{\textbf{DVIS++ + POSReasoner}} & R50
        & \textbf{56.0} & 80.6 & \textbf{60.9}
        & \textbf{51.1} & \textbf{72.6} & \textbf{55.8}
        & \textbf{47.0} & \textbf{68.2} & \textbf{51.0} \\
        \addlinespace[1.5pt]
        DAQ & R50
        & 54.6 & 78.2 & 60.3
        & 50.1 & 71.6 & 55.1
        & 45.9 & 66.7 & 50.0 \\
        \rowcolor{POSRbg}
        \textcolor{POSRaccent}{\textbf{DAQ + POSReasoner}} & R50
        & \textbf{55.5} & \textbf{79.2} & \textbf{60.9}
        & \textbf{51.0} & \textbf{72.9} & \textbf{55.9}
        & \textbf{46.7} & \textbf{67.9} & \textbf{50.9} \\
        \addlinespace[1.5pt]
        LOMM$^\dagger$ & R50
        & 55.8 & 80.1 & 61.5
        & 50.9 & 73.3 & 56.6
        & 47.2 & 68.8 & 51.9 \\
        \rowcolor{POSRbg}
        \textcolor{POSRaccent}{\textbf{LOMM + POSReasoner}}$^\dagger$ & R50
        & \textbf{56.1} & 80.0 & \textbf{62.1}
        & \textbf{51.3} & \textbf{73.6} & \textbf{56.9}
        & \textbf{47.7} & \textbf{69.2} & \textbf{52.4} \\
        \addlinespace[1.5pt]
        LOMM & ViT-L
        & 69.1 & 89.3 & 76.5
        & 65.0 & 85.8 & 72.7
        & 60.4 & 82.7 & 67.1 \\
        \rowcolor{POSRbg}
        \textcolor{POSRaccent}{\textbf{LOMM + POSReasoner}} & ViT-L
        & \textbf{69.8} & \textbf{89.5} & \textbf{77.0}
        & \textbf{65.8} & \textbf{86.0} & \textbf{73.3}
        & \textbf{61.4} & \textbf{83.4} & \textbf{67.6} \\
        \bottomrule
    \end{tabularx}
    \endgroup
\end{table}

\paragraph{Reading the extended comparisons.}
Tables~\ref{tab:ovis_full_preview} and~\ref{tab:ytvis_full_preview} provide
complementary views of the evaluation. Cited rows place the tested systems
within the published benchmark landscape, whereas adjacent host and
POSReasoner rows isolate the contribution of the added reasoner. Each pair
shares the host checkpoint, candidate masks, input resolution, and evaluator;
only POSReasoner is trained. The appropriate reference for a plug-in gain is
therefore the paired host, rather than another published implementation of
the same architecture. This distinction is particularly relevant to
YouTube-VIS 2022, whose published comparison rows follow LOMM. The offline
refinement and provisional-result markers are retained so that the evaluation
setting remains explicit for every row.

On OVIS, all five evaluated host--backbone pairs improve in AP, AP50, and
AP75. With ResNet-50, the AP gains are 2.7 points for CTVIS, 2.1 for DVIS++,
and 1.6 for DAQ. These hosts use different temporal designs, yet each benefits
from reasoning over persistent states. Moreover, each of these OVIS gains
exceeds the corresponding host's gain on any of the three YouTube-VIS
versions. This pattern is consistent with the motivation for retaining
identity evidence in crowded videos: when current observations are ambiguous,
state history supplies a reference that can be revisited before an association
is committed. The benefit is thus observed across the evaluated hosts, not
only with one association architecture.

The DAQ results further separate the choice of visual backbone from the
addition of state reasoning. AP rises from 38.3 to 39.9 with ResNet-50,
from 49.6 to 50.6 with Swin-L, and from 53.9 to 55.2 with ViT-L.
These paired improvements show that the reasoner remains useful across
different feature representations and host performance levels. The gains
also persist at the stricter AP75 threshold: DAQ improves by 1.8, 1.1, and
1.4 points, respectively. These results are consistent with a complementary
benefit from state reasoning across the evaluated backbones: a higher host
score does not eliminate the observed improvement from adding POSReasoner.

Across YouTube-VIS 2019, 2021, and 2022, AP and AP75 improve in all 15
reported host--version comparisons, including the provisionally marked
LOMM/ResNet-50 configuration. CTVIS gains 1.0, 1.1, and 2.1 AP across the
three versions, while DAQ gains 0.9, 0.9, and 0.8. LOMM with ViT-L also
improves on every version, by 0.7, 0.8, and 1.0 AP. The repeated direction
of these paired changes supports the applicability of the state interface
across the evaluated VIS systems. Comparing the versions separately also
preserves their distinct evaluation settings, rather than combining their
scores into a single aggregate.

AP50 and AP75 help interpret these results alongside overall AP. AP75
requires closer mask overlap, and its improvements show that the benefit is
visible beyond the more permissive AP50 threshold. Since the candidate masks
remain fixed within each pair, the comparison concerns how available
predictions are selected, associated, and carried through time, rather than
training a new mask generator. Video-instance AP jointly reflects these
decisions and the resulting track quality. The per-threshold columns provide
complementary measurements, while the state and component analyses examine
the temporal behavior more directly.

In particular, the controls in Table~\ref{tab:vis_reasoning_appendix} connect
the benchmark improvements to identity-specific history and structured
interaction. Host cues alone and shuffled states remain near the frozen
host; intact states without a graph reach 36.55 AP, and the full model
reaches 37.31 AP. The LVOS study in Section~\ref{sec:ablation} separately
examines retention, reactivation, and cross-trajectory reasoning, while
Appendix~\ref{sec:reappearance_analysis} groups recovery events by absence
duration. Together, these comparisons give the extended tables a behavioral
context: the benchmark pairs establish where the reasoner helps, and the
targeted analyses examine the roles of persistent history, renewed
observations, and competing identities within that reasoning process.

\end{document}